\pdfoutput=1

\documentclass[11pt]{article}

\usepackage[final]{acl}

\usepackage{times}
\usepackage{latexsym}

\usepackage[T1]{fontenc}

\usepackage[utf8]{inputenc}

\usepackage{microtype}

\usepackage{inconsolata}

\usepackage{graphicx}

\usepackage{amsmath}
\usepackage{amssymb}
\usepackage{amsfonts}
\usepackage{booktabs}
\usepackage{multicol}
\usepackage{multirow}

\newcommand{\phdiff}[1]{\!\colorbox[cmyk]{0.3,0,0.3,0}{#1}\!}
\newcommand{\pldiff}[1]{\!\colorbox[cmyk]{0.1,0,0.1,0}{#1}\!}
\newcommand{\nhdiff}[1]{\!\colorbox[cmyk]{0,0.3,0.3,0}{#1}\!}
\newcommand{\nldiff}[1]{\!\colorbox[cmyk]{0,0.1,0.1,0}{#1}\!}

\definecolor{green_mention}{HTML}{38761D}
\definecolor{orange_mention}{HTML}{f29637}
\definecolor{gray_mention}{HTML}{666666}

\usepackage{CJKutf8}
\usepackage{subcaption}
\usepackage{fontawesome}

\title{Disambiguating Reference in Visually Grounded Dialogues through\\Joint Modeling of Textual and Multimodal Semantic Structures}

\author{
 \textbf{Shun Inadumi \textsuperscript{1,2}},
 \textbf{Nobuhiro Ueda \textsuperscript{3,*}},
 \textbf{Koichiro Yoshino \textsuperscript{4,2,1}},
\\
 \textsuperscript{1} Nara Institute of Science and Technology,
 \textsuperscript{2} Guardian Robot Project, RIKEN \\
 \textsuperscript{3} Kyoto University,
 \textsuperscript{4} Institute of Science Tokyo
\\
 \texttt{inazumi.shun.in6@naist.ac.jp} \\
 \texttt{ueda@nlp.ist.i.kyoto-u.ac.jp}
 \texttt{koichiro@c.titech.ac.jp}
}

\begin{document}
\maketitle
\renewcommand{\thefootnote}{\fnsymbol{footnote}}  
\footnotetext[1]{Currently at NEC Corporation.}
\renewcommand{\thefootnote}{\arabic{footnote}}  

\begin{abstract}
Multimodal reference resolution, including phrase grounding, aims to understand the semantic relations between mentions and real-world objects.
Phrase grounding between images and their captions is a well-established task. 
In contrast, for real-world applications, it is essential to integrate textual and multimodal reference resolution to unravel the reference relations within dialogue, especially in handling ambiguities caused by pronouns and ellipses.
This paper presents a framework that unifies textual and multimodal reference resolution by mapping mention embeddings to object embeddings and selecting mentions or objects based on their similarity.\footnote{The code is publicly available at \url{https://github.com/SInadumi/mmrr}.}
Our experiments show that learning textual reference resolution, such as coreference resolution and predicate-argument structure analysis, positively affects performance in multimodal reference resolution.
In particular, our model with coreference resolution performs better in pronoun phrase grounding than representative models for this task, MDETR and GLIP.
Our qualitative analysis demonstrates that incorporating textual reference relations strengthens the confidence scores between mentions, including pronouns and predicates, and objects, which can reduce the ambiguities that arise in visually grounded dialogues.
\end{abstract}

\section{Introduction}

\begin{figure}[t]
    \centering
    \includegraphics[keepaspectratio, scale=0.19]{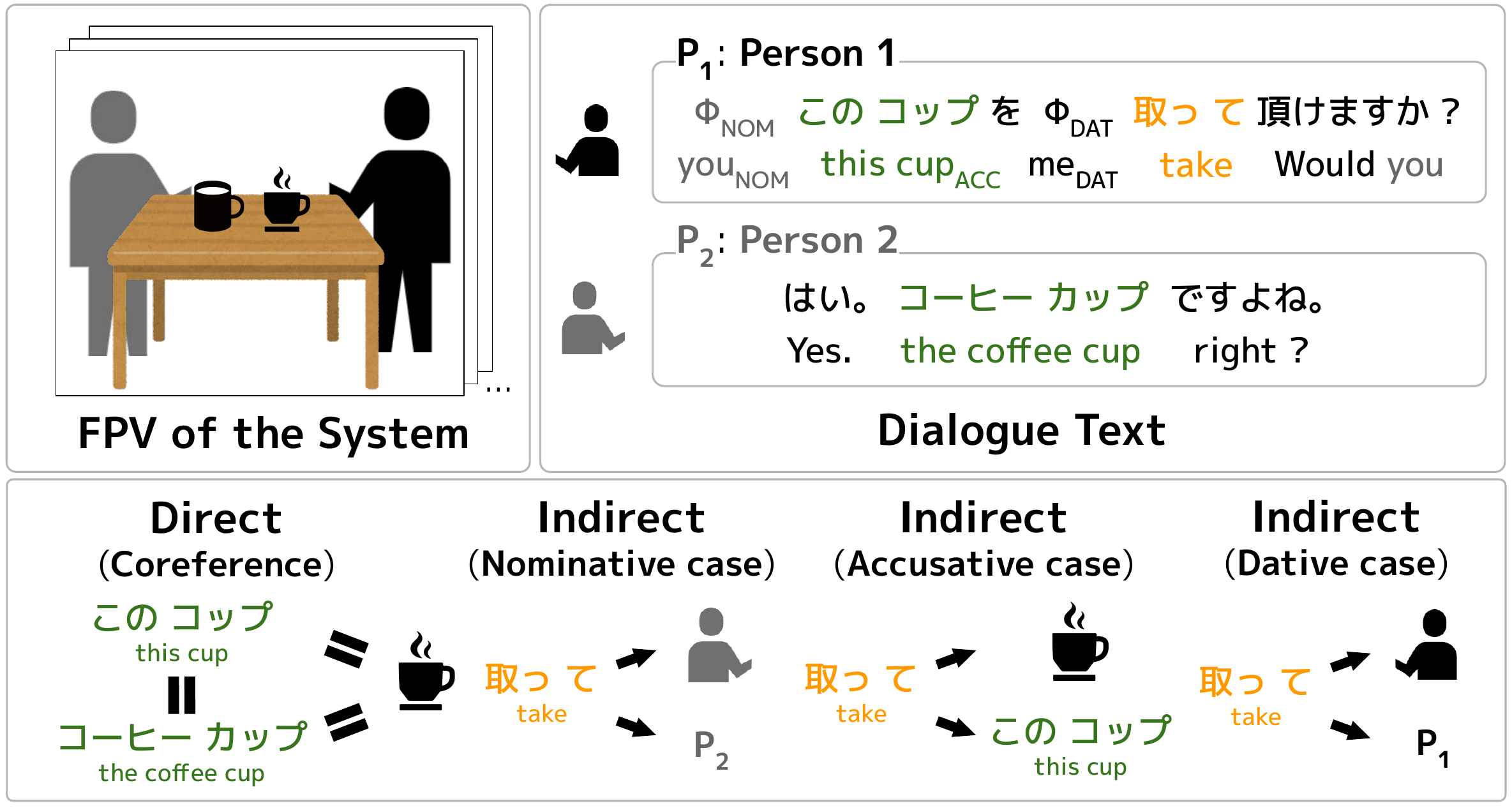}
    \caption{Example of textual and multimodal reference resolutions in the system analyzes a two-person dialogue, ``Would you take me this cup? --- Yes. the coffee cup, right?,'' from its first-person view. Japanese omits the \textbf{\textcolor{gray_mention}{subject}} and \textbf{object} of the \textbf{\textcolor{orange_mention}{predicate}} ``take.''}
    \label{fig:1_top}
\end{figure}

Understanding what mentions refer to objects in visually grounded dialogues is key to realizing a system that can collaborate with users in the real world, including robots and embodied agents~\cite{yu-etal-2019-see, kottur-etal-2021-simmc, wu-etal-2023-simmc, ueda-etal-2024-j}.
Recent studies have focused on identifying the objects referred to by mentions, as exemplified by ``the coffee cup'' and ``this cup'' in Figure~\ref{fig:1_top}, which are called direct references.

Among these studies, multimodal reference resolution is a task that identifies semantic structures~\cite{fillmore1967case,clark-1975-bridging}, not only direct references but also indirect references between mentions and objects~\cite{ueda-etal-2024-j}.
Figure~\ref{fig:1_top} illustrates how the system understands the objects referred to indirectly by ``take.''
In this example, the system can identify the objects referred to by ``this cup'' and ``the coffee cup'' through both direct and indirect references, inferred from the predicate-argument structures of ``take.'' 
This allows the system to understand ``who does what to whom,'' ``when,'' and ``where'' in a dialogue by relating events to objects.

Pronouns and ellipses are typical instances of direct and indirect reference. 
These are frequently used in dialogue and pose challenges for multimodal reference resolution~\cite{ueda-etal-2024-j}.
Phrase grounding between images and their captions is a well-established task~\cite{kazemzadeh-etal-2014-referitgame, flickrentitiesijcv}; however, identifying direct references is still insufficient in visually grounded dialogues.\footnote{Example of Table~\ref{tab:4_quantitative_results_coref-to-phrase_grounding} shows the performance of GLIP~\cite{Li_2022_CVPR} for phrase grounding in Japanese dialogue. The Recall@1 is 0.377, significantly worse than that for general Japanese captions.}
Figure~\ref{fig:1_top} illustrates an example where pronouns and ellipses of topical terms, such as subjects and objects, present difficulties for such a framework.

To address these challenges, we aim to improve the performance of multimodal reference resolution, thereby disambiguating references in visually grounded dialogues.
Previous work on phrase grounding in dialogues~\cite{das-etal-2017-visdial} shows that coreference resolution improves pronoun grounding performance~\cite{yu-etal-2022-vd-pcr,lu-etal-2022-extending}.
Furthermore, joint modeling of textual references\footnote{It is a general term that encompasses coreference, case relations~\cite{fillmore1967case} in predicate-argument structures, and bridging anaphora~\cite{clark-1975-bridging}.} can improve performance in textual reference resolution tasks~\cite{shibata-kurohashi-2018-entity, omori-komachi-2019-multi, ueda-etal-2020-bert}.
Inspired by these works, we hypothesize that introducing linguistic features of textual references can also benefit multimodal reference resolution in dialogues.
As ``this cup'' and ``the coffee cup'' refer to the same coffee cup, resolving the coreference between the mentions may help determine the direct and indirect reference.

In this study, we propose a framework for the joint modeling of all references.
Our framework unifies textual reference resolution --- including coreference resolution and predicate-argument structure analysis~\cite{iida-etal-2007-annotating} --- and multimodal reference resolution, by following recent advances in multimodal representation learning~\cite{gupta-etal-2020-ECCV, pmlr-v139-radford21a, Li_2022_CVPR}.
We map mention embeddings to object embeddings and select mentions or objects based on their similarity.
Our multimodal reference resolution model explicitly addresses indirect references and handles ellipses.

Quantitative results using our framework show the effectiveness of training textual reference relations in improving the performance in analyzing direct ({\S}~\ref{sec:4_results_phrase_grounding}) and indirect ({\S}~\ref{sec:4_results_reference_resolution}) references.
Our findings suggest that textual reference resolution positively contributes to multimodal reference resolution.
In particular, our model with coreference resolution performs better in pronoun phrase grounding than representative models for this task, MDETR~\cite{Kamath_2021_ICCV} and GLIP~\cite{Li_2022_CVPR}.

Our qualitative analysis shows that incorporating textual reference relations strengthens the confidence scores between mentions, including pronouns and predicates, and objects.
These findings are also consistent with our quantitative results.
For example, in Figure~\ref{fig:1_top}, we observe an increase in the confidence scores for predicting the objects referred to by mentions such as ``this cup'' and ``take.''
Thus, this study demonstrates that textual reference resolution can reduce the ambiguities in visually grounded dialogues, particularly those caused by pronouns and ellipses.

\section{Preliminaries}
A dataset for multimodal reference resolution in two-party dialogues, the J-CRe3\footnote{J-CRe3: Japanese Conversation Dataset for Real-world Reference Resolution}~\cite{ueda-etal-2024-j}, which this study employs for experiments.
The following describes reference resolution ({\S}~\ref{sec:2_reference_resolution}), which consists of textual reference resolution (TRR, {\S}~\ref{sec:2_textual_reference_resolution}) and multimodal reference resolution (MRR, {\S}~\ref{sec:2_multimodal_reference_resolution}).

\subsection{Task Settings}
\subsubsection{Reference Resolution}
\label{sec:2_reference_resolution}
Given a text $\mathbf{T}$ and a sequence of images $\mathbf{V}=\{\mathbf{I}_1  \cdots \mathbf{I} \cdots\}$ corresponding to $\mathbf{T}$, reference resolution identifies the reference relations that exist between mentions and objects or events they refer to.
This task consists of TRR, which analyzes between mentions, and MRR, which analyzes between mentions and objects.

As shown in Figure~\ref{fig:1_top}, many instances of reference relations connected by direct reference are clarified by the chain of cases between the predicate and its arguments. 
In addition to direct references (marked by ``=''), we define five labels (i.e., case) to represent the types of semantic connections in indirect references (Table~\ref{tab:2_mapping_reference_to_anaphora}).
Let $L$ denote the set of six types of all reference relations $l$.

\begin{table}[t]
    \centering
    \scalebox{0.77}{
        \begin{tabular}{c|c}
            \toprule
             \multirow{2}{*}{\textbf{Japanese case marker}} & \textbf{Case and anaphora relations} \\
             & (\textbf{Abbreviations}) \\
             \cmidrule(lr){0-1}
             ``\textit{ga}'' (\begin{CJK}{UTF8}{ipxm}ガ\end{CJK}) & Nominative case (\textbf{NOM})   \\
             ``\textit{wo}'' (\begin{CJK}{UTF8}{ipxm}ヲ\end{CJK}) & Accusative case (\textbf{ACC})   \\
             ``\textit{ni}'' (\begin{CJK}{UTF8}{ipxm}ニ\end{CJK}) & Dative case (\textbf{DAT})   \\
             \multirow{2}{*}{``\textit{de}'' (\begin{CJK}{UTF8}{ipxm}デ\end{CJK})} & Instrumental case (\textbf{INS}) \\
             & Locative case (\textbf{LOC}) \\
             ``\textit{no}'' (\begin{CJK}{UTF8}{ipxm}ノ\end{CJK}) & Bridging anaphora \\
             \bottomrule
        \end{tabular}
    }
    \caption{Types of indirect references: We show the corresponding cases~\cite{fillmore1967case} and bridging anaphora~\cite{clark-1975-bridging} based on Japanese case markers.}
    \label{tab:2_mapping_reference_to_anaphora}
\end{table}

\subsubsection{Textual Reference Resolution}
\label{sec:2_textual_reference_resolution}
Given a text $\mathbf{T}$, TRR identifies phrases that have a reference relation with another phrase. 
We refer to such phrases as mentions.
We use the term TRR to refer collectively to coreference resolution, predicate-argument structure (PAS) analysis~\cite{iida-etal-2007-annotating}, and bridging anaphora (BA) resolution~\cite{poesio-vieira-1998-corpus,kobayashi-ng-2020-bridging,yu-poesio-2020-multitask}.

\subsubsection{Multimodal Reference Resolution}
\label{sec:2_multimodal_reference_resolution}
Given a text $\mathbf{T}$ and an image $\mathbf{I}$, MRR identifies objects in $\mathbf{I}$ that have reference relations with mentions in $\mathbf{T}$.
Specifically, MRR involves an object detection process that estimates up to $q$ tuples of bounding boxes $\mathbf{O}$ and object feature series $\mathbf{X}$ for $\mathbf{I}$, denoted as $(\mathbf{O},\mathbf{X}) =\{(o_1, \mathbf{x}_1) \cdots (o_q, \mathbf{x}_q)\}$.
Based on $\mathbf{T}$ and $\mathbf{X}$, we identify elements in $\mathbf{O}$ that have reference relations with a mention.
Identifying only direct references (``='') to objects from a mention is called phrase grounding~\cite{kazemzadeh-etal-2014-referitgame, flickrentitiesijcv}.

\subsection{J-CRe3 Annotation}
\label{sec:2_jcre3}
J-CRe3~\cite{ueda-etal-2024-j} is a dataset of real-world interactions during collaborative work between a master and an assistive robot, including first-person video of the robot, third-person video, and audio/transcription of their dialogue.
We use first-person videos, dialogue transcriptions, and annotations for reference resolution, including textual reference relations, mention-to-object direct/indirect reference relations, and bounding boxes.

Ambiguous expressions, including pronouns and ellipses, frequently occur in spoken language~\cite{PIANTADOSI2012280}, with ellipses occurring especially in Japanese~\cite{seki-etal-2002-probabilistic}, Chinese~\cite{kong-zhou-2010-tree}, and Korean~\cite{park-etal-2015-zero}.
In the ellipses, for example, indirect references from predicates to objects can exist without explicitly mentioning those objects.
Following the previous work~\cite{ueda-etal-2024-j}, we refer to these instances as zero references, as in the case of zero anaphora in text~\cite{sasano-etal-2008-fully}.
While datasets addressing pronouns have existed in the past~\cite{das-etal-2017-visdial, kottur-etal-2021-simmc, wu-etal-2023-simmc, Goel_2023_ICCV}, J-CRe3 differs in that it also includes zero references.\footnote{Although the work of~\citet{oguz-etal-2023-find,oguz-etal-2024-mmar} addresses noun phrase ellipsis, J-CRe3 is the only dataset that provides annotations for the explicit estimation of zero references from predicates to objects.}


\begin{figure*}[t]
    \centering
    \includegraphics[keepaspectratio, scale=0.222]{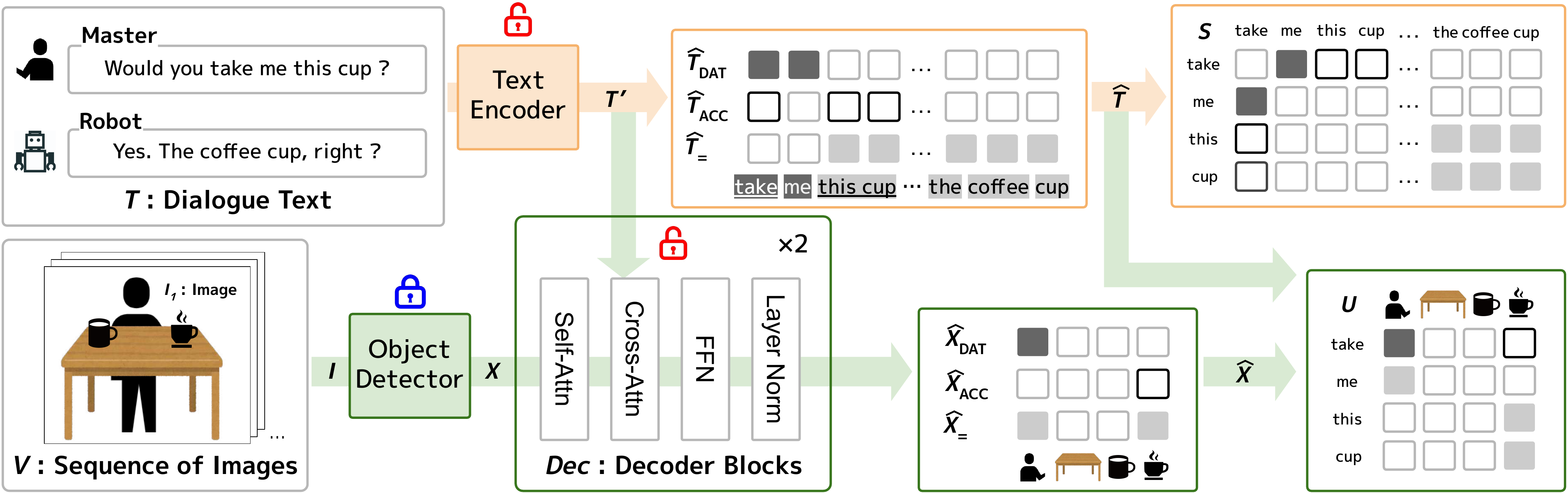}
    \caption{Overview of our framework for J-CRe3: The orange and green indicate the processing flows for TRR and MRR, respectively. }
    \label{fig:3_framework}
\end{figure*}

\section{Methodology}

\subsection{Motivation}
Visually grounded dialogue datasets, including J-CRe3, have limited training data, especially in Japanese and minor languages.
For this reason, state-of-the-art models such as MDETR~\cite{Kamath_2021_ICCV} and GLIP~\cite{Li_2022_CVPR}, which require large image-text pairs for training, are difficult to train on Japanese datasets only.

The previous work proposed a method for handling MRR that treats phrase grounding and TRR independently before combining their results~\cite{ueda-etal-2024-j}.
However, this method is unable to account for zero references.

To address these problems, we design an MRR model inspired by weakly supervised phrase grounding models~\cite{gupta-etal-2020-ECCV,goel-etal-2023-semi}, which can be trained with limited data.
Our unified framework integrates learning and analysis processes for TRR and MRR using mention and object embeddings.
In particular, our MRR model explicitly considers indirect references between mentions and objects, with reference to the existing Japanese text analyzer~\cite{ueda-etal-2023-kwja}.

The difficulty of MRR lies in the need for the system to also recognize references for pronouns and ellipses.
In our integrated framework, we expect TRR to enhance MRR by supplementing information for ambiguous direct and indirect references, such as from pronouns and ellipses.

\subsection{Overview of Our Unified Framework}
Figure~\ref{fig:3_framework} shows the overview of our unified framework for reference resolution.
Our MRR model uses a frozen object detector~\cite{zhou2022detic} while fine-tuning a text encoder~\cite{devlin-etal-2019-bert} and a fusion module~\cite{Liu_2024_ECCV} that integrates text $\mathbf{T}$ with an object feature series $\mathbf{X}$.
We train two models separately: one for TRR ({\S}~\ref{sec:3_TRR_model}) and the other for MRR ({\S}~\ref{sec:3_MRR_model}). 
They share the text encoder weights.

\subsubsection{Textual Reference Resolution Model}
\label{sec:3_TRR_model}
Our TRR model for analyzing all textual reference relations in Figure~\ref{fig:3_framework} is based on the similarity of embeddings between mentions.

Given a dialogue text $\mathbf{T}$, the text encoder outputs subword embeddings $\mathbf{T}'\in \mathbb{R}^{p \times d_T}$ of input length $p$ and dimension $d_T$.
The extended representation $\widehat{\mathbf{T}}$ for $\mathbf{T}'$ is as follows:
\begin{align}
    \label{form:3_1}
    \widehat{\mathbf{T}} &= \mathbf{T}' \mathbf{W}_{T1} \in \mathbb{R}^{p \times d_T \times |L|},
\end{align}
where, $\mathbf{W}_{T1} \in \mathbb{R}^{d_T \times d_T \times |L|}$ is trainable parameters.
Our model calculates the dot product of each $\widehat{\mathbf{T}}$ per a relation $l \in L$ and the similarity matrix $\mathbf{S}_{l}$ as follows:
\begin{align}
    \label{form:3_2}
    \mathbf{S}_{l} &= \widehat{\mathbf{T}}_l {\widehat{\mathbf{T}}_{l}^{\top}} \in \mathbb{R}^{p \times p}.
\end{align}
We use $\mathbf{S}_{l}$ to select a mention that the other mention refers to.

The embeddings $\mathbf{T}'$ are at a subword level, but a mention is at a basic phrase level.\footnote{This phrase consists of one content word and zero or more function words.}
To link these different units, we use the first subword of a mention as its main representation during learning and inference.
This step is the same as for our MRR model described below.

\subsubsection{Multimodal Reference Resolution Model}
\label{sec:3_MRR_model}
Our MRR model for analyzing direct and indirect reference relations in Figure~\ref{fig:3_framework} is based on the similarity of mention and object embeddings.

Given a dialogue text $\mathbf{T}$ and an image $\mathbf{I}$, which is an element of frames $\mathbf{V}$, a text encoder and an object detector output $\mathbf{T}'$ and $(\mathbf{O},\mathbf{X})$, respectively.
Then, our model uses a single linear layer and aligns the dimension $d_T$ of $\mathbf{T}'$ and the dimension $d_O$ of $\mathbf{X}$ to $d_S$.
The extended representation $\widehat{\mathbf{T}}$ and $\widehat{\mathbf{X}}$ for $\mathbf{T}'$ and $\mathbf{X}$ as follows:
\begin{align}
    \label{form:3_3}
    \widehat{\mathbf{T}} &= \mathbf{T}' \mathbf{W}_{T2} \in \mathbb{R}^{ p \times d_S \times |L|}, \\
    \label{form:3_4}
    \widehat{\mathbf{X}} &= Dec(\mathbf{X}, \mathbf{T}') \mathbf{W}_O \in \mathbb{R}^{ q \times d_S \times |L|}, \\
    & \quad \ Dec(\mathbf{X}, \mathbf{T}') \in \mathbb{R}^{q \times d_S} \nonumber,
\end{align}
where $(\mathbf{W}_{T2}, \mathbf{W}_O) \in \mathbb{R}^{d_S \times d_S \times |L|}$ are trainable parameters and $Dec(\cdot)$ is two decoder blocks, which uses cross-attention to condition $\mathbf{X}'$ on $\mathbf{T}'$~\cite{Liu_2024_ECCV}.
Our model calculates the dot product of each $\widehat{\mathbf{T}}$ and $\widehat{\mathbf{X}}$ per a relation $l \in L$ and the similarity matrix $\mathbf{U}_l$ as follows:
\begin{align}
    \label{form:3_5}
    \mathbf{U}_l &= \widehat{\mathbf{X}}_{l} \widehat{\mathbf{T}}_{l}^{ \top} \in \mathbb{R}^{p \times q}.
\end{align}
We use the similarity matrix $\mathbf{U}_{l}$ to select elements of $\mathbf{O}$ from a mention.

We consider frames $\mathbf{V}$ to be a sequence of one-second intervals extracted from a video from the start and end times of an utterance in a text $\mathbf{T}$.
Following previous work~\cite{gupta-etal-2020-ECCV,goel-etal-2023-semi}, we use pooled features~\cite{Lin_2017_CVPR,Anderson_2018_CVPR} from the region proposal network~\cite{NIPS2015_14bfa6bb} as object feature series $\mathbf{X}$.

\subsubsection{Loss Functions}
Using the softmax cross entropy used on phrase grounding~\cite{Li_2022_CVPR} and Japanese text analysis~\cite{ueda-etal-2020-bert}, we define the loss functions for as follows:
\begin{align}
    \label{form:3_6} 
    \mathcal{L}_{S} &= \sum_{l\in L} loss\{ \mathbf{S}_l; \mathbf{S}_{(l, ground)}\}, \\
    \label{form:3_7} 
    \mathcal{L}_{U} &= \sum_{l\in L} loss\{ \mathbf{U}_l; \mathbf{U}_{(l, ground)}\},
\end{align}
where $\mathcal{L}_{S}$ corresponds to TRR model, and $\mathcal{L}_{U}$ corresponds to MRR model.
Here, $\mathbf{S}_{(l, ground)} \in \{0,1\}^{p \times p}$ and $\mathbf{U}_{(l, ground)} \in \{0,1\}^{p \times q}$ represent matrices of positive examples of $\mathbf{S}_l$ and $\mathbf{U}_l$ in a reference relation $l$.

\section{Experiments}

\subsection{Settings}
\paragraph{Compared Models}
We assume the MRR-only model is a baseline model ({\S}~\ref{sec:3_MRR_model}). 
In our experiments, we first train a TRR model by only coreference resolution, predicate-argument structure (PAS) analysis and bridging anaphora (BA) resolution, or TRR.
Then, we leverage this text encoder to train the baseline model. 
This approach allows us to investigate how TRR benefits phrase grounding and MRR.
As TRR models, we use our TRR model ({\S}~\ref{sec:3_TRR_model}) and a Japanese text analyzer, KWJA~\cite{ueda-etal-2023-kwja}.
Both models can handle TRR well, but this study focuses on results in MRR.

For phrase grounding comparison with the MRR models, we fine-tune MDETR~\cite{Kamath_2021_ICCV} and GLIP~\cite{Li_2022_CVPR}, representative models for this task on Japanese datasets.
For MRR comparison, we consider a method that combines phrase grounding outputs from GLIP with TRR outputs from KWJA~\cite{ueda-etal-2024-j}, denoted as GLIP + KWJA.
Specifically, GLIP performs phrase grounding on both text and image and outputs only i) direct reference relations, while KWJA performs TRR on text only and outputs ii) all textual reference relations.
We derive indirect reference relations by aligning the i) and ii) relations.

\paragraph{Dataset}
We use J-CRe3 and Flickr30k-Ent-JP~\cite{nakayama-etal-2020-visually} as Japanese datasets to fine-tune our MRR model, MDETR, and GLIP.
J-CRe3 contains 93 dialogues and a total of 11,062 images, with each dialogue containing 10 to 16 utterances.
Flickr30k-Ent-JP includes images and corresponding Japanese captions with direct references between mentions and objects, totaling 31,783 captions and 63,566 images.

We pre-train MDETR and GLIP using Visual Genome~\cite{ranjay_2017_vis_genome}, GQA~\cite{Hudson_2019_CVPR}, and Flickr30k-Ent-JP, with these weights serving as initial values for fine-tuning on the Japanese datasets. 
In contrast, the MRR models do not undergo pre-training and are instead fine-tuned directly on these datasets.

To train TRR models, we use J-CRe3 as well as a corpus of web documents~\cite{hangyo-etal-2012-building}, Wikipedia, and blog posts annotated with textual reference relations.
These datasets contain 6,542 documents and dialogues for the training set.

\paragraph{Evaluation Metrics}
We use Recall@\textit{k} (R@\textit{k}; $k=\{1,5,10\}$) to evaluate phrase grounding and MRR. 
An MRR model, MDETR, and GLIP predict bounding boxes and their confidence scores for each mention.  
Recall@\textit{k} is the percentage of times that ground truth boxes are among the top \textit{k} predicted boxes with the highest confidence scores. 
We consider predicted boxes to match ground truth boxes if their Intersection-over-Union is 0.5 or greater.

\begin{table}[t]
\centering
    \scalebox{0.68}{
        \begin{tabular}{l|cccc}
        \toprule
        \textbf{Models} & \textbf{Text Encoder} & \textbf{Object Detector} & \textbf{Others} & \textbf{Total} \\
        \cmidrule(lr){0-4}
        Ours & 339M & --- & 27M & 366M \\
        GLIP & 278M & 31M & 92M & 401M \\
        MDETR & 278M & 8M & 20M & 306M \\
        \bottomrule
        \end{tabular}
    }
    \caption{Number of trainable parameters in the models: While the total number of parameters in our models remains mostly unchanged between MRR and phrase grounding, task-specific components such as $W_{T2}$ and $W_{O}$ differ depending on the task.}
    \label{tab4:model_parameters}
\end{table}

\paragraph{Implementation Details}
Our framework uses Japanese DeBERTa-v2-large~\cite{he2021deberta} as a text encoder, and Detic~\cite{zhou2022detic} with Swin-Transformer~\cite{Liu_2021_ICCV} as its backbone for an object detector. 
MDETR and GLIP use mDeBERTa-v3-base~\cite{he2021debertav3} as a text encoder.
Table~\ref{tab4:model_parameters} shows the number of trainable parameters for the MRR models developed in our framework, GLIP, and MDETR.

We set the maximum length of the subword embeddings $\mathbf{T}'$ to $p=256$ and the dimension to be $d_T=d_O=d_S=1,024$.
The MRR models output the maximum value of the predicted bounding boxes $O$ for the two datasets: $q=128$ for the J-CRe3 and $q=256$ for the Flickr30k-Ent-JP.

For the TRR, each training instance consists of three sentences, shifting by one sentence at a time.
For phrase grounding and MRR, the instance unit definitions vary by dataset. 
In J-CRe3, each training instance comprises three utterances, shifting by one utterance at a time, paired with an image; evaluation is performed on individual utterance–image pairs.\footnote{See Appendix~\ref{sec:appendix_ablation_window-size} for ablation study results on the utterance length of $\mathbf{T}$ during evaluation of phrase grounding and MRR.}
In Flickr30k-Ent-JP, each instance, used for training and evaluation, consists of up to five captions paired with an image.

We fine-tuned the TRR and MRR models using AdamW~\cite{loshchilov2018decoupled} with a learning rate of 5e-5, weight decay of 0.01, and 1,000 warmup steps and trained for 16 epochs with a batch size of 16 and 32. 
MDETR and GLIP are also fine-tuned with the same settings, except for 2 epochs and a batch size of 4 and 16.
We performed the TRR and MRR models experiments with 4$\times$RTX 3090s in 6 hours and MDETR and GLIP experiments with 2$\times$RTX A6000s in 2 days.

\begin{table*}[t]
    \begin{minipage}[t]{0.67\linewidth}
    \centering
    \scalebox{0.68}{
    \begin{tabular}{l|ccc|ccc|ccc}
        \toprule
         \multirow{2}{*}{\textbf{Models}} & \multicolumn{3}{c}{\textbf{Overall} (996)} & \multicolumn{3}{c}{\textbf{Nouns} (671 / 996)} & \multicolumn{3}{c}{\textbf{Pronouns} (120 / 996)} \\
         & R@1 & R@5 & R@10 & R@1 & R@5 & R@10 & R@1 & R@5 & R@10 \\
         \cmidrule(lr){0-9}
         \multicolumn{10}{c}{\textbf{Coreference Resolution} (Coref.) $\Rightarrow$ \textbf{Phrase Grounding} (PG)} \\
         Baseline $^{\dagger}$ & 0.342 & 0.567 & 0.649 & 0.344 & 0.574 & 0.664 & 0.277 & 0.527 & 0.641 \\
         \ w/ KWJA $^{\dagger}$ & \nldiff{0.315} & \nldiff{0.542} & \nldiff{0.640} & \nldiff{0.308} & \nldiff{0.549} & \nldiff{0.657} & \pldiff{0.283} & 0.527 & \nldiff{0.633} \\
         \ w/ Ours $^{\dagger}$ & \pldiff{0.348} & \pldiff{0.584} & \pldiff{0.681} & \nldiff{0.339} & \nldiff{0.567} & 0.663 & \phdiff{\textbf{0.361}} & \phdiff{\textbf{0.683}} & \phdiff{\textbf{0.772}}  \\
         \cmidrule(lr){0-9}
         GLIP & \textbf{0.445} & \textbf{0.745} & \textbf{0.808} & \textbf{0.454} & \textbf{0.752} & \textbf{0.816} & 0.241 & 0.650 & 0.733 \\
         MDETR & 0.348  & 0.468 & 0.510 & 0.365 & 0.481 & 0.523 & 0.133 & 0.208 & 0.250 \\ 
         \bottomrule
    \end{tabular}
    }
    \end{minipage}
    \hspace{0.01\linewidth}
    \begin{minipage}[t]{0.28\linewidth}
    \scalebox{0.68}{
        \begin{tabular}{l|ccc}
        \toprule
        \textbf{Models} & R@1 & R@5 & R@10\\
        \cmidrule(lr){0-3}
        \multicolumn{4}{c}{\textbf{Coref.} $\Rightarrow$ \textbf{PG}}\\
        Baseline $^{\dagger}$ & 0.558  & 0.735 & 0.767 \\
        \ w/ KWJA $^{\dagger}$ & 0.559 & 0.733 & 0.767 \\
        \ w/ Ours $^{\dagger}$ & 0.560 & 0.733 & 0.767 \\
        \cmidrule(lr){0-3}
        GLIP & \textbf{0.822} & \textbf{0.951} & \textbf{0.970} \\
        MDETR & 0.769 & 0.896 & 0.924 \\
        \bottomrule
        \end{tabular}
    }
    \end{minipage}
    \caption{The results of phrase grounding: For models with $\dagger$, we report the average of 3 randomly seeded training and evaluation iterations of MRR. The highlighted items indicate where the MRR models with coreference resolution show \pldiff{improvements} or \nldiff{deteriorations} compared to the baseline. \textbf{Left}: The results of J-CRe3: The numbers in parentheses indicate positive instances. \textbf{Right}: The results of Flickr30k-Ent-JP.}
    \label{tab:4_quantitative_results_coref-to-phrase_grounding}
\end{table*}
\begin{figure*}[t]
    \centering
    \includegraphics[keepaspectratio, scale=0.28]{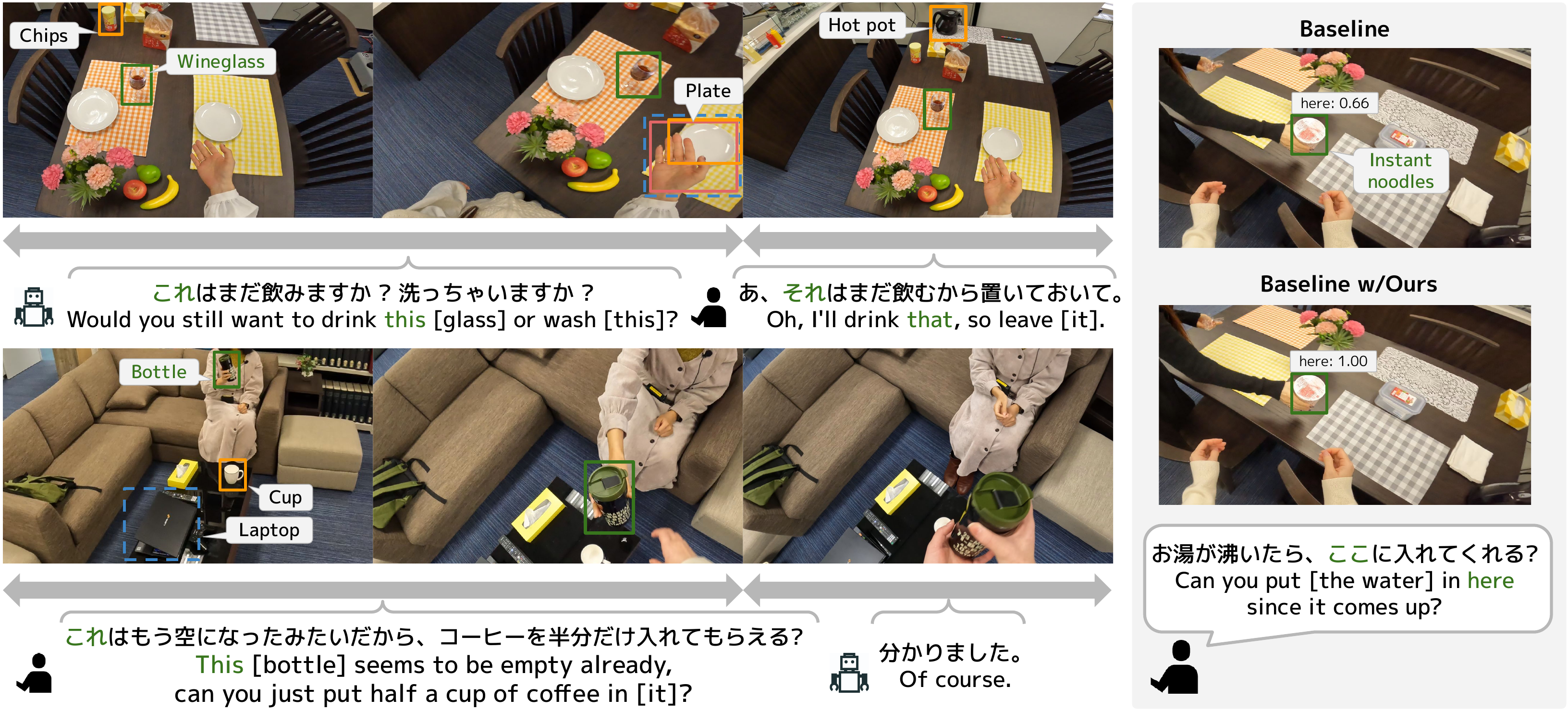}
    \caption{Examples of phrase grounding: The green mentions and objects are targets for grounding, and the mentions in square brackets are omitted in Japanese. \textbf{Left}: We show the Recall@1 errors of models in orange (GLIP), blue dashed (Baseline), and red (Baseline w/ Ours), while only incorrectly predicted bounding boxes are shown, as correct predictions are omitted. \textbf{Right}: We also show confidence scores.}
    \label{fig:4_examples_grounding}
\end{figure*}

\subsection{Experiments on Phrase Grounding}
\label{sec:4_results_phrase_grounding}

\paragraph{Main Results}
Table~\ref{tab:4_quantitative_results_coref-to-phrase_grounding} shows the results of phrase grounding.
In the overall evaluation of J-CRe3, including noun phrases and pronouns, our MRR model with coreference resolution using our TRR model (Baseline w/ Ours) outperforms the baseline and MDETR in terms of Recall@5 and 10.
While our MRR model performs slightly worse than the baseline for nouns, it considerably outperforms all other models, including GLIP, for pronouns.

Based on J-CRe3 results, we compare the baseline, Baseline w/ Ours, and an MRR model using KWJA as a TRR model (Baseline w/ KWJA).
When our TRR model is used for phrase grounding, it improves performance for pronouns while minimizing performance degradation for nouns compared to KWJA.
Our unified framework highlights the effectiveness of coreference resolution in phrase grounding for visually grounded dialogues.

\paragraph{Performance on Flickr30k-Ent-JP}
In the evaluation on Flickr30k-Ent-JP, GLIP shows the highest performance, and no change in Recall@\textit{k} due to coreference resolution was observed for either Baseline w/ Ours or Baseline w/ KWJA.
Unlike MDETR and GLIP, the MRR models use a frozen object detector, including Baseline w/ Ours.
Thus, the upper bound of Recall@\textit{k} for object detection depends on the detector. 
The actual upper bound is 0.799, which is the limiting value of Recall@\textit{k} for MRR models.

\begin{figure}[t]
    \centering
    \includegraphics[keepaspectratio, scale=0.525]{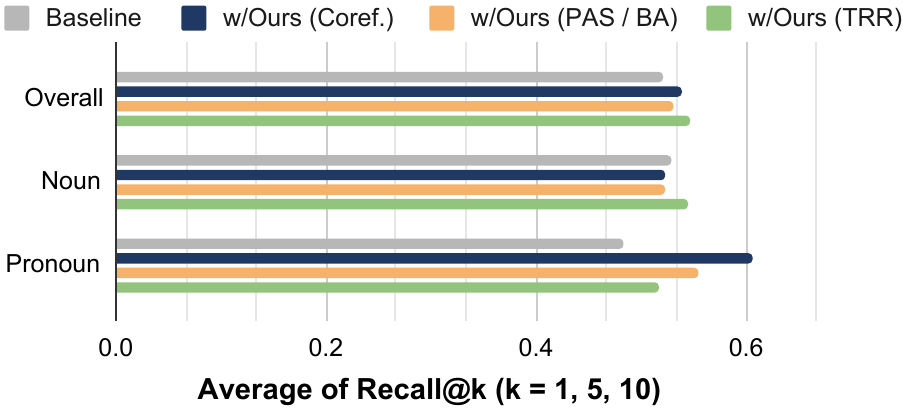}
    \caption{Ablation study results of Baseline w/ Ours in phrase grounding: We compare the improvements achieved by coreference resolution (Coref.),  predicate-argument structure analysis and bridging anaphora resolution (PAS / BA), and textual reference resolution (TRR). See Table~\ref{tab:appendix_phrase_grounding_details} for detailed results.}
    \label{fig:4_ablation_phrase_grounding}
\end{figure}

\begin{table*}[t]
    \centering
    \scalebox{0.68}{
        \begin{tabular}{l|ccc|ccc|ccc|ccc|ccc}
        \toprule
         \multirow{2}{*}{\textbf{Models}} & \multicolumn{3}{c}{\textbf{NOM} (2,053)} & \multicolumn{3}{c}{\textbf{ACC} (915)} & \multicolumn{3}{c}{\textbf{DAT} (1,074)} & \multicolumn{3}{c}{\textbf{INS}--\textbf{LOC} (139)} & \multicolumn{3}{c}{\textbf{Bridging} (163)} \\
        & R@1 & R@5 & R@10 & R@1 & R@5 & R@10 & R@1 & R@5 & R@10 & R@1 & R@5 & R@10 & R@1 & R@5 & R@10 \\
         \cmidrule(lr){0-15}
         \multicolumn{16}{c}{\textbf{Textual Reference Resolution} (TRR) $\Rightarrow$ \textbf{Multimodal Reference Resolution} (MRR)} \\
         Baseline $^{\dagger}$ & 0.568 & 0.735 & 0.763 & 0.229 & 0.505 & 0.606 & 0.559 & 0.726 & 0.749 & 0.124 & 0.350 & 0.465 & 0.378 & 0.564 & 0.662 \\
         \ w/ KWJA $^{\dagger}$  & \pldiff{0.574} & \pldiff{\textbf{0.756}} & \pldiff{\textbf{0.785}} & \pldiff{\textbf{0.240}} & 0.506 & \nldiff{0.601} & \pldiff{\textbf{0.582}} & \pldiff{\textbf{0.748}} & \pldiff{\textbf{0.779}} & \phdiff{0.199} & \phdiff{0.431} & \phdiff{\textbf{0.561}} & \pldiff{\textbf{0.411}} & \pldiff{\textbf{0.613}} & \pldiff{\textbf{0.680}} \\
         \ w/ Ours $^{\dagger}$  & \pldiff{\textbf{0.585}} & \pldiff{0.745} & \pldiff{0.773} & 0.230 & \pldiff{\textbf{0.520}} & 0.607 & \pldiff{0.576} & \pldiff{0.735} & \pldiff{0.772} & \pldiff{0.172} & \phdiff{0.424} & \phdiff{0.532} & \pldiff{0.386} & \pldiff{0.588} & 0.662 \\
         \cmidrule(lr){0-15}
         GLIP + KWJA & 0.060 & 0.111 & 0.118 & 0.190 & 0.386 & 0.420 & 0.065 & 0.079 & 0.081 & \textbf{0.273} & \textbf{0.510} & 0.539 & 0.226 & 0.288 & 0.294 \\
         \bottomrule
        \end{tabular}
    }
    \caption{The results of indirect references in MRR on the J-CRe3: The highlighted items indicate where the MRR models with TRR show \pldiff{improvements} or \nldiff{deteriorations} compared to the baseline. See the caption of Figure~\ref{tab:4_quantitative_results_coref-to-phrase_grounding} for $\dagger$ and parentheses.}
    \label{tab:4_quantitative_results_in_mmref-indirect}
\end{table*}

\paragraph{Qualitative Analysis}
Figure~\ref{fig:4_examples_grounding} shows the examples of phrase grounding results in Table~\ref{tab:4_quantitative_results_coref-to-phrase_grounding}.
As shown on the left side in Figure~\ref{fig:4_examples_grounding}, our MRR model has fewer Recall@1 errors for pronouns such as ``this'' (``\begin{CJK}{UTF8}{ipxm}これ\end{CJK}'') and ``it'' (``\begin{CJK}{UTF8}{ipxm}それ\end{CJK}'') compared to the baseline and GLIP.
Since all models, including our MRR model, rely on inference from a single image, they exhibited inconsistencies in predictions. 
For example, some errors involved estimating an object like ``plate'' for a mention of ``this.''

The right side of Figure~\ref{fig:4_examples_grounding} shows the examples of the baseline and our MRR model.
Both models accurately predict the pronoun ``here'' (``\begin{CJK}{UTF8}{ipxm}ここ\end{CJK}''), but their confidence scores for the object ``instant noodle'' differ: the baseline assigned a score of 0.66, whereas our model assigned a score of 1.00. 
Thus, our qualitative analysis demonstrates that incorporating coreference relations strengthens the confidence of pronoun-to-object predictions.

\paragraph{Comparison of Textual Reference Relations}
We discuss the benefits of incorporating textual reference relations into phrase grounding using the baseline and our MRR model. 
Figure~\ref{fig:4_ablation_phrase_grounding} shows the ablation results for MRR models with coreference resolution, PAS analysis and BA resolution or TRR, in phrase grounding.

Incorporating textual reference relations improves pronoun performance compared to the baseline, regardless of the relation type, with coreference relations being the most effective. 
Furthermore, incorporating all textual reference relations yields the best performance for nouns. 
These results demonstrate the benefits of jointly modeling direct references, coreference, and case and bridging anaphora references, in phrase grounding for visually grounded dialogues.

\begin{figure*}[t]
    \centering
    \includegraphics[keepaspectratio, scale=0.31]{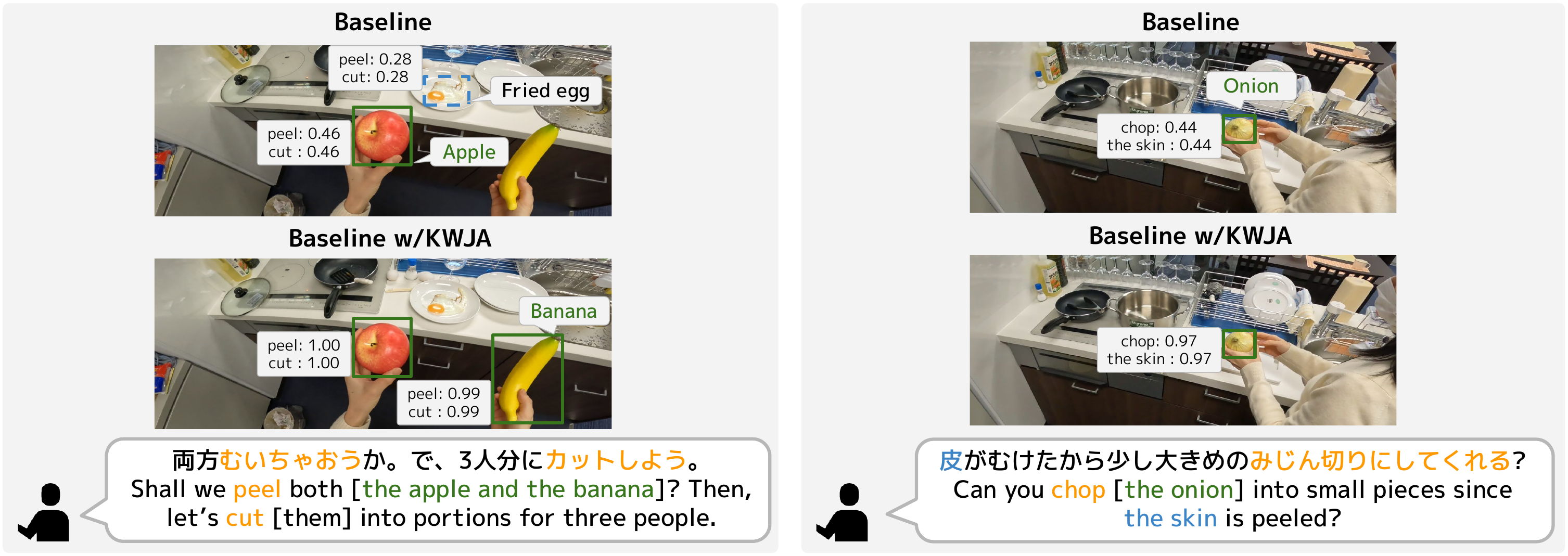}
    \caption{Examples of indirect references in MRR: The orange and blue mentions are targets for the accusative case and bridging anaphora, respectively. The green mentions correspond to referring objects. See the caption of Figure~\ref{fig:4_examples_grounding} for square brackets and green and blue dashed boxes.}
    \label{fig:4_examples_mmref}
\end{figure*}

\subsection{Experiments on Multimodal Reference Resolution}
\label{sec:4_results_reference_resolution}
\paragraph{Main Results}
Table~\ref{tab:4_quantitative_results_in_mmref-indirect} shows the results of indirect references in MRR.
Our MRR model with TRR (Baseline w/ Ours) consistently outperforms the baseline and GLIP + KWJA in terms of Recall@10, regardless of the TRR models. 
However, GLIP + KWJA achieves the highest Recall@\textit{1} and 5 for instrumental and locative cases (INS-LOC). 
Since GLIP + KWJA cannot handle zero references, the performance of INS-LOC in J-CRe3 probably depends on phrase grounding performance.

\paragraph{Qualitative Analysis}
Figure~\ref{fig:4_examples_mmref} shows the examples of indirect reference results in MRR of Table~\ref{tab:4_quantitative_results_in_mmref-indirect}.
Here, we focus on the Baseline w/ KWJA, which performs well in Table~\ref{tab:4_quantitative_results_in_mmref-indirect}.

The left side of Figure~\ref{fig:4_examples_mmref} shows an example of zero references with two objects, ``the apple'' and ``the banana,'' which are targets of accusative case relations (ACC), referred to by the mentions ``peel'' (``\begin{CJK}{UTF8}{ipxm}むいちゃおう\end{CJK}'') and ``cut'' (``\begin{CJK}{UTF8}{ipxm}カットしよう\end{CJK}''). 
Compared to the baseline, the Baseline w/ KWJA correctly analyzes these objects.

The right side of Figure~\ref{fig:4_examples_mmref} shows an ACC and a bridging anaphora as examples of indirect reference results.
Both models correctly analyzed the referenced object (``onion'') from mentions, but the confidence scores were higher for Baseline w/ KWJA.
Our qualitative analysis suggests that textual reference relations strengthen the prediction of objects for predicates and anaphoric mentions in learning MRR.

\begin{figure}[t]
    \centering
    \includegraphics[keepaspectratio, scale=0.51]{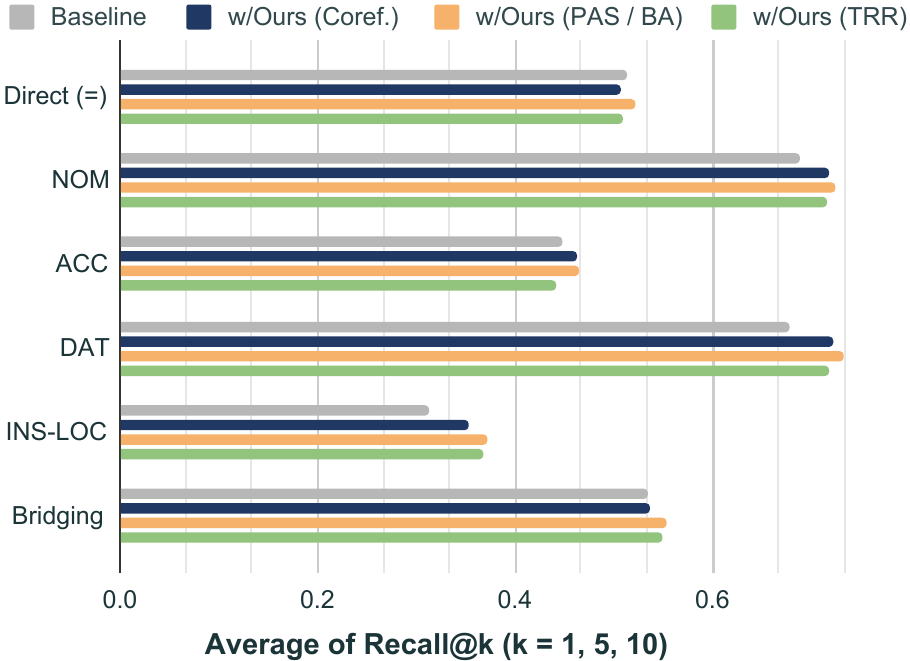}
    \caption{Ablation study results of Baseline w/ Ours in MRR: See the caption of Figure~\ref{fig:4_ablation_phrase_grounding} for a detailed comparison setting and Tables~\ref{tab:appendix_mmref-indirect_details} and~\ref{tab:appendix_mmref-direct_details} for detailed results.}
    \label{fig:4_ablation_mmref}
\end{figure}

\paragraph{Comparison of Textual Reference Relations}
We discuss the benefits of incorporating textual reference relations into MRR using the baseline and our MRR model.
Figure~\ref{fig:4_ablation_mmref} shows the ablation results for MRR models with coreference resolution, PAS analysis and BA resolution or TRR in MRR.

The results for direct references in MRR show that our model did not outperform the baseline, regardless of whether coreference resolution, PAS analysis and BA resolution, or TRR were included.
We speculate that this is because the expressive power of our model was primarily allocated to analyzing indirect references in MRR.

Our model, with PAS analysis and BA resolution, achieved the best performance on the results for indirect references in MRR, although coreferences sometimes hindered its performance.
Thus, incorporating such textual reference relations is beneficial for analyzing indirect references in the MRR of dialogues.

\section{Related Work and Discussion}
Existing phrase grounding models can be broadly divided into two architectures:
\begin{itemize}
    \item [i)] Freeze the weights of the object detector and use the bounding boxes of the detection results as pseudo-labels, often called weakly supervised phrase grounding~\cite{anna-etal-2016-ECCV, Datta_2019_ICCV, wang-etal-2020-maf, gupta-etal-2020-ECCV, goel-etal-2023-semi}.
    \item [ii)] Incorporate object detection into the training model and dynamically detect boxes according to input text~\cite{Kamath_2021_ICCV, Li_2022_CVPR, Liu_2024_ECCV}.
\end{itemize}
Our MRR model follows the approach i) to reduce the learning costs, while we chose the approach ii) as the comparison model for their higher phrase grounding performance.

Previous studies have explored the benefits of coreference and anaphora resolution for pronoun phrase grounding in English~\cite{yu-etal-2022-vd-pcr, lu-etal-2022-extending, oguz-etal-2023-find} and multilingual settings~\cite{oguz-etal-2024-mmar}, focusing on dialogue and procedural texts. 
These findings are partially consistent with our results on phrase grounding ({\S}~\ref{sec:4_results_phrase_grounding}).
However, our study goes further by comprehensively investigating indirect references between mentions and objects ({\S}~\ref{sec:4_results_reference_resolution}) to apply these findings to real-world systems, such as assistive robots.
The ability to identify the object referred to by a predicate is crucial for task planning in collaborative robots, as it directly contributes to their success rate~\cite{toki-and-heannette-2022-grounding, shirai-etal-2022-vision, han2024interpret}. 
Our findings are, therefore, naturally related to this field.

Our proposed framework provides an approach to MRR by explicitly handling zero references, a capability lacking in methods that combine separate phrase grounding and TRR models~\cite{ueda-etal-2024-j}, such as GLIP + KWJA used in our experiment. 
While recent vision-and-language models based on large language models~\cite{hurst2024gpt} can potentially achieve comparable analysis through prompting techniques~\cite{yang2023setofmark}, their application to videos such as J-CRe3 remains challenging due to the time-consuming and high computational costs.
\section{Conclusion}
This paper has presented a unified framework for textual and multimodal reference resolution to disambiguate references in visually grounded dialogues. 
Our results showed that incorporating textual reference relations improved performance in multimodal reference resolution, including phrase grounding.
In particular, our model with coreference resolution outperformed representative models on phrase grounding for pronouns.
Further analysis demonstrated that incorporating textual reference relations strengthens the confidence scores for pronouns, predicates, and anaphoric mentions of objects.
In future research, we plan to explore the cross-lingual applicability of our framework to languages other than Japanese.
Furthermore, we will improve the multimodal reference resolution model by augmenting the multimodal references and applying this to assistive robots.

\section*{Acknowledgments}
This work was supported by RIKEN Junior Research Associate Program and JST, PRESTO Grant Number JPMJPR24TC, Japan.

\section*{Limitations}
\paragraph{Data}
We acknowledge that our experimental results are limited to two-party Japanese dialogues. 
Specifically, for indirect references, the nominative and dative case objects referred to by anaphoric mentions are usually either the master or the robot (See {\S}~\ref{sec:2_jcre3}). 
Future research should explore other languages and multi-party dialogue settings to investigate the applicability of our framework beyond these constraints.

\paragraph{Models}
Our MRR model is constrained by the frozen object detector, which results in lower Recall@\textit{k} scores for noun phrase grounding compared to GLIP. 
Additionally, all models rely on inference from a single image, which makes it challenging to maintain prediction consistency across different visual contexts.

To address these issues, future research will first focus on integrating the object detection process to enhance the MRR model performance further. 
This improvement will involve data augmentation techniques, such as leveraging large language models to generate visually grounded dialogues and reference relations by combining existing images or videos. 
Moreover, exploring video-based architectures and incorporating a first-person view of the system are expected to improve prediction consistency and resolve ambiguities in visually grounded dialogues by leveraging user movements across sequential frames.

\paragraph{Experiments}
Further analysis --- including a detailed analysis of textual indirect references --- is needed to fully understand the effect of TRR on MRR.
Moreover, investigating the correlation between TRR and MRR performance could provide valuable insights, though this would require multiple training and evaluation iterations of TRR and MRR models.
\section*{Ethical Consideration}
This study primarily utilized publicly available datasets, such as Flickr30k-Ent-JP and J-CRe3, to prevent ethical concerns.
However, J-CRe3 contains videos of identifiable individuals who participated in the data collection process.
Therefore, to safeguard their privacy, any use of models trained on J-CRe3 should be used with caution, particularly when intended for commercial use.

\bibliography{custom}

\appendix
\section{Resources}
\begin{flushleft}
\subsection{Data}
\begin{itemize}
    \item \textbf{Flickr30kEnt-JP}~\cite{nakayama-etal-2020-visually}:  \url{https://github.com/nlab-mpg/Flickr30kEnt-JP}.
    \item \textbf{J-CRe3}~\cite{ueda-etal-2024-j}: \url{https://github.com/riken-grp/J-CRe3}.
    \item \textbf{Visual Genome}~\cite{ranjay_2017_vis_genome}: \url{https://homes.cs.washington.edu/~ranjay/visualgenome}.
    \item \textbf{GQA}~\cite{Hudson_2019_CVPR}: \url{https://cs.stanford.edu/people/dorarad/gqa}.
    \item \textbf{Kyoto University Web Document Leads Corpus}~\cite{hangyo-etal-2012-building}: \url{https://github.com/ku-nlp/KWDLC}.
    \item \textbf{Wikipedia Annotated Corpus}: \url{https://github.com/ku-nlp/WikipediaAnnotatedCorpus}.
    \item \textbf{Annotated FKC Corpus}: \url{https://github.com/ku-nlp/AnnotatedFKCCorpus}.
\end{itemize}

\subsection{Model}
\begin{itemize}
    \item \textbf{Japanese DeBERTa-v2-large}~\cite{he2021deberta}: \url{https://huggingface.co/ku-nlp/deberta-v2-large-japanese}.
    \item \textbf{mDeBERTa-v3-base}~\cite{he2021debertav3}: \url{https://huggingface.co/microsoft/mdeberta-v3-base}.
    \item \textbf{Detic}~\cite{zhou2022detic}: \url{https://github.com/facebookresearch/Detic}; we used \url{Detic_LCOCOI21k_CLIP_SwinB_896b32_4x_ft4x_max-size} as a frozen object detector.
    \item \textbf{MDETR}~\cite{Kamath_2021_ICCV}: \url{https://github.com/ashkamath/mdetr}.
    \item \textbf{GLIP}~\cite{Li_2022_CVPR}: \url{https://github.com/microsoft/GLIP}.
\end{itemize}

\subsection{Software}
\begin{itemize}
    \item \textbf{KWJA}~\cite{ueda-etal-2023-kwja}: \url{https://github.com/ku-nlp/kwja}.
    \item \textbf{Multi-modal Reference Resolution}~\cite{ueda-etal-2024-j}: \url{https://github.com/riken-grp/multimodal-reference}.
\end{itemize}
\end{flushleft}
\section{Detailed Results in Multimodal Reference Resolution}
We present detailed quantitative evaluation results for phrase grounding and MRR using the MRR models in Tables~\ref{tab:appendix_phrase_grounding_details},~\ref{tab:appendix_mmref-indirect_details}, and~\ref{tab:appendix_mmref-direct_details}.

\begin{table*}[p]
    \centering
    \scalebox{0.68}{
    \begin{tabular}{l|c|ccc|ccc|ccc}
        \toprule
         \multicolumn{2}{c}{\multirow{2}{*}{\textbf{Models}}} & \multicolumn{3}{c}{\textbf{Overall} (996)} & \multicolumn{3}{c}{\textbf{Nouns} (671 / 996)} & \multicolumn{3}{c}{\textbf{Pronouns} (120 / 996)} \\
         \multicolumn{2}{c}{} & R@1 & R@5 & R@10 & R@1 & R@5 & R@10 & R@1 & R@5 & R@10 \\
         \cmidrule(lr){0-10}
         \multirow{2}{*}{Baseline} & $\mu$ & 0.342 & 0.567 & 0.649 & \underline{0.344} & 0.574 & 0.664 & 0.277 & 0.527 & 0.641 \\
         & $\pm \sigma$ & 0.008 & 0.011 & 0.023 & 0.006 & 0.014 & 0.005 & 0.020 & 0.091 & 0.118\\ 
         \cmidrule(lr){0-10}
         Baseline & \multicolumn{10}{c}{\textbf{Coreference Resolution} $\Rightarrow$ \textbf{Phrase Grounding}} \\
         \multirow{2}{*}{\ w/ KWJA} & $\mu$ & \nldiff{0.315} & \nldiff{0.542} & \nldiff{0.640} & \nldiff{0.308} & \nldiff{0.549} & \nldiff{0.657} & \pldiff{0.283} & 0.527 & \nldiff{0.633} \\
         & $\pm \sigma$ & 0.024 & 0.008 & 0.031 & 0.029 & 0.009 & 0.011 & 0.036 & 0.054 & 0.094\\
         \multirow{2}{*}{\ w/ Ours} & $\mu$ & \pldiff{\underline{0.348}}  & \pldiff{0.584} & \pldiff{0.681} & \nldiff{0.339} & \nldiff{0.567} & 0.663 & \phdiff{\textbf{0.361}} & \phdiff{\textbf{0.683}} & \phdiff{\textbf{0.772}}  \\
         & $\pm \sigma$ & 0.020 & 0.028 & 0.016 & 0.029 & 0.023 & 0.010 & 0.070 & 0.071 & 0.050 \\
         \cmidrule(lr){0-10}
         Baseline & \multicolumn{10}{c}{\textbf{PAS Analysis} and \textbf{BA Resolution} $\Rightarrow$ \textbf{Phrase Grounding}} \\
         \multirow{2}{*}{\ w/ KWJA} & $\mu$ & \nhdiff{0.258} & \nhdiff{0.476} & \nhdiff{0.566} & \nhdiff{0.265} & \nhdiff{0.485} & \nhdiff{0.575} & \nldiff{0.230} & \nhdiff{0.491} & \nhdiff{0.594} \\
         & $\pm \sigma$ & 0.115 & 0.141 & 0.122 & 0.135 & 0.158 & 0.151 & 0.070 & 0.136 & 0.067 \\
         \multirow{2}{*}{\ w/ Ours} & $\mu$ & 0.338 & \pldiff{0.580} & \pldiff{0.672} & \nldiff{0.329} & 0.571 & 0.667 & \pldiff{0.311} & \phdiff{0.613} & \phdiff{0.738} \\
         & $\pm \sigma$ & 0.014 & 0.014 & 0.020 & 0.016 & 0.024 & 0.014 & 0.026 & 0.048 & 0.048 \\
         \cmidrule(lr){0-10}
          Baseline & \multicolumn{10}{c}{\textbf{Textual Reference Resolution} $\Rightarrow$ \textbf{Phrase Grounding}} \\
         \multirow{2}{*}{\ w/ KWJA} & $\mu$ & \nldiff{0.325} & \nldiff{0.549} & \nldiff{0.627} & 0.340 & 0.570 & \nldiff{0.656} & \pldiff{0.302} & \pldiff{0.550} & \nldiff{0.597} \\
         & $\pm \sigma$ & 0.026 & 0.026 & 0.036 & 0.009 & 0.005 & 0.004 & 0.047 & 0.038 & 0.050 \\
         \multirow{2}{*}{\ w/ Ours} & $\mu$ & \pldiff{\underline{0.347}} & \pldiff{\textbf{0.600}} & \pldiff{\textbf{0.689}} & \underline{0.345} & \pldiff{\textbf{0.597}} & \pldiff{\textbf{0.690}} & 0.258 & \phdiff{0.597} & \phdiff{0.694} \\
         & $\pm \sigma$ & 0.016 & 0.016 & 0.013 & 0.021 & 0.038 & 0.020 & 0.036 & 0.037 & 0.066 \\
         \bottomrule
    \end{tabular}
    }
    \caption{Detail results of phrase grounding on the J-CRe3: We report the average ($\mu$) and standard deviation ($\pm \sigma$) of 3 randomly seeded training and evaluation iterations.}
    \label{tab:appendix_phrase_grounding_details}
\end{table*}

\begin{table*}[p]
    \centering
    \scalebox{0.66}{
        \begin{tabular}{l|c|ccc|ccc|ccc|ccc|ccc}
        \toprule
         \multicolumn{2}{c}{\multirow{2}{*}{\textbf{Models}}} & \multicolumn{3}{c}{\textbf{NOM} (2,053)} & \multicolumn{3}{c}{\textbf{ACC} (915)} & \multicolumn{3}{c}{\textbf{DAT} (1,074)} & \multicolumn{3}{c}{\textbf{INS}--\textbf{LOC} (139)} & \multicolumn{3}{c}{\textbf{Bridging} (163)} \\
         \multicolumn{2}{c}{} & R@1 & R@5 & R@10 & R@1 & R@5 & R@10 & R@1 & R@5 & R@10 & R@1 & R@5 & R@10 & R@1 & R@5 & R@10 \\
         \cmidrule(lr){0-16}
         Baseline & $\mu$ & 0.568 & 0.735 & 0.763 & 0.229 & 0.505 & 0.606 & 0.559 & 0.726 & 0.749 & 0.124 & 0.350 & 0.465 & 0.378 & 0.564 & 0.662 \\
         & $\pm \sigma$  & 0.014 & 0.030 & 0.034 & 0.007 & 0.027 & 0.017 & 0.008 & 0.031 & 0.037 & 0.039 & 0.047 & 0.004 & 0.028 & 0.053 & 0.061 \\
         \cmidrule(lr){0-16}
         Baseline & \multicolumn{16}{c}{\textbf{Coreference Resolution} $\Rightarrow$ \textbf{Multimodal Reference Resolution}} \\
         \multirow{2}{*}{\ w/ KWJA} & $\mu$ & \nldiff{0.542} & \nldiff{0.697} & \nldiff{0.726} & \pldiff{0.261} & \nldiff{0.492} & \nldiff{0.583} & \pldiff{0.576} & 0.725 & \pldiff{0.753} & \phdiff{0.177} & \phdiff{0.414} & \phdiff{0.537} & \pldiff{0.390} & \pldiff{0.578} & \nldiff{0.650} \\
         & $\pm \sigma$  & 0.052 & 0.124 & 0.144 & 0.010 & 0.052 & 0.096 & 0.054 & 0.101 & 0.121 & 0.018 & 0.008 & 0.020 & 0.039 & 0.044 & 0.090 \\
         \multirow{2}{*}{\ w/ Ours} & $\mu$ & \pldiff{0.572} & \pldiff{0.772} & \pldiff{0.812} & \pldiff{\textbf{0.267}} & \pldiff{\textbf{0.527}} & \pldiff{\underline{0.618}} & \pldiff{\underline{0.585}} & \phdiff{0.779} & \phdiff{0.818} & \pldiff{0.146} & \pldiff{0.400} & \phdiff{0.520} & \nldiff{0.372} & 0.564 & \pldiff{0.689} \\
         & $\pm \sigma$  & 0.011 & 0.002 & 0.010 & 0.017 & 0.013 & 0.012 & 0.014 & 0.001 & 0.007  & 0.014 & 0.010 & 0.018 & 0.019 & 0.021 & 0.030 \\
         \cmidrule(lr){0-16}
         Baseline & \multicolumn{16}{c}{\textbf{PAS Analysis} and \textbf{BA Resolution} $\Rightarrow$ \textbf{Multimodal Reference Resolution}} \\
         \multirow{2}{*}{\ w/ KWJA} & $\mu$ & \nldiff{0.549} & 0.732 & 0.761 & \pldiff{0.236} & \nldiff{0.497} & \nldiff{0.587} & \pldiff{0.574} & \pldiff{0.727} & \pldiff{0.752} & \phdiff{0.203} & \pldiff{0.376} & \pldiff{0.477} & \nldiff{0.370} & 0.560 & 0.644 \\
         & $\pm \sigma$  & 0.009 & 0.027 & 0.039 & 0.010 & 0.008 & 0.029 & 0.009 & 0.028 & 0.034 & 0.046 & 0.043 & 0.023 & 0.009 & 0.014 & 0.040 \\
         \multirow{2}{*}{\ w/ Ours} & $\mu$ & \pldiff{0.575} & \pldiff{\textbf{0.776}} & \phdiff{\textbf{0.822}} & \pldiff{0.253} & \pldiff{0.519} & \pldiff{\underline{0.620}} & \pldiff{\underline{0.585}} & \phdiff{\textbf{0.784}} & \phdiff{\textbf{0.829}} & \phdiff{0.194} & \phdiff{0.410} & \pldiff{0.508} & \nldiff{0.370} & \pldiff{0.595} & \pldiff{\textbf{0.697}} \\
         & $\pm \sigma$  & 0.013 & 0.003 & 0.007 & 0.026 & 0.021 & 0.010 & 0.009 & 0.009 & 0.003 & 0.031 & 0.007  & 0.004 & 0.033 & 0.037 & 0.030 \\
         \cmidrule(lr){0-16}
         Baseline & \multicolumn{16}{c}{\textbf{Textual Reference Resolution} $\Rightarrow$ \textbf{Multimodal Reference Resolution}} \\
         \multirow{2}{*}{\ w/ KWJA} & $\mu$ & \pldiff{0.574} & \pldiff{0.756} & \pldiff{0.785} & \pldiff{0.240} & 0.506 & \nldiff{0.601} & \pldiff{\underline{0.582}} & \pldiff{0.748} & \pldiff{0.779} & \phdiff{\textbf{0.199}} & \phdiff{\textbf{0.431}} & \phdiff{\textbf{0.561}} & \pldiff{\textbf{0.411}} & \pldiff{\textbf{0.613}} & \pldiff{0.680} \\
         & $\pm \sigma$  & 0.005 & 0.006 & 0.006 & 0.014 & 0.008 & 0.018 & 0.027 & 0.012 & 0.015 & 0.025 & 0.044 & 0.073 & 0.022 & 0.016 & 0.018 \\
         \multirow{2}{*}{\ w/ Ours} & $\mu$ & \pldiff{\textbf{0.585}} & \pldiff{0.745} & \pldiff{0.773} & 0.230 & \pldiff{0.520} & 0.607 & \pldiff{0.576} & \pldiff{0.735} & \pldiff{0.772} & \pldiff{0.172} & \phdiff{0.424} & \phdiff{0.532} & \pldiff{0.386} & \pldiff{0.588} & 0.662 \\
         & $\pm \sigma$  & 0.006 & 0.017 & 0.023  & 0.013 & 0.032 & 0.034 & 0.009 & 0.024 & 0.028 & 0.004 & 0.025 & 0.007 & 0.023 & 0.049 & 0.026 \\
         \bottomrule
        \end{tabular}
    }
    \caption{Detail results of indirect references in MRR on the J-CRe3: See the caption of Table~\ref{tab:appendix_phrase_grounding_details} for $\mu$ and $\pm \sigma$.}
    \label{tab:appendix_mmref-indirect_details}
\end{table*}

\begin{table*}[p]
    \centering
    \scalebox{0.68}{
        \begin{tabular}{l|c|ccc|ccc|ccc}
            \toprule
            \multicolumn{2}{c}{\multirow{2}{*}{\textbf{Models}}} & \multicolumn{3}{c}{\textbf{Overall} (996)} & \multicolumn{3}{c}{\textbf{Nouns} (671 / 996)} & \multicolumn{3}{c}{\textbf{Pronouns} (120 / 996)} \\
            \multicolumn{2}{c}{} & R@1 & R@5 & R@10 & R@1 & R@5 & R@10 & R@1 & R@5 & R@10  \\
             \cmidrule(lr){0-10}
             \multirow{2}{*}{Baseline} & $\mu$ & 0.313 & \underline{0.564} & 0.658 & 0.304 & 0.552 & 0.643 & 0.252 & 0.566 & 0.697 \\
             & $\pm \sigma$ & 0.013 & 0.026 & 0.011 & 0.016 & 0.036 & 0.022 & 0.012 & 0.014 & 0.026 \\
             \cmidrule(lr){0-10}
             Baseline &  \multicolumn{10}{c}{\textbf{Coreference Resolution} $\Rightarrow$ \textbf{Multimodal Reference Resolution}} \\
             \multirow{2}{*}{\ w/ KWJA} & $\mu$ & \pldiff{0.325} & \nldiff{0.556} & \nldiff{0.642} & \pldiff{\underline{0.335}} & \pldiff{\underline{0.569}} & \pldiff{\underline{0.653}} & \phdiff{\textbf{0.316}} & \pldiff{\textbf{0.583}} & \nldiff{0.691} \\
             & $\pm \sigma$ & 0.018  & 0.034 & 0.072 & 0.046 & 0.007 & 0.031 & 0.072 & 0.068 & 0.101 \\
             \multirow{2}{*}{\ w/ Ours} & $\mu$ & \pldiff{0.319} & \nldiff{0.549} & \nldiff{0.650} & \pldiff{\underline{0.334}} & 0.550 & \nldiff{0.636} & \nldiff{0.244} & \nldiff{0.558} & \pldiff{\textbf{0.708}} \\
             & $\pm \sigma$ & 0.017 & 0.014 & 0.006 & 0.011 & 0.005 & 0.009 & 0.020 & 0.008 & 0.014 \\
             \cmidrule(lr){0-10}
             Baseline &  \multicolumn{10}{c}{\textbf{PAS Analysis} and \textbf{BA Resolution} $\Rightarrow$ \textbf{Multimodal Reference Resolution}} \\
             \multirow{2}{*}{\ w/ KWJA} & $\mu$ & \nldiff{0.295} & \nldiff{0.534} & \nldiff{0.624} & 0.304 & \nldiff{0.545} & \nldiff{0.634} & \pldiff{0.280} & \nhdiff{0.513} & \nhdiff{0.636} \\
             & $\pm \sigma$ & 0.016 & 0.028 & 0.046 & 0.014 & 0.010 & 0.025 & 0.025 & 0.026 & 0.029 \\
             \multirow{2}{*}{\ w/ Ours} & $\mu$ & \pldiff{\textbf{0.331}} & \underline{0.563} & \pldiff{\textbf{0.667}} & \pldiff{0.315} & \nldiff{0.542} & 0.646 & \pldiff{0.277} & \nldiff{0.558} & 0.694 \\
             & $\pm \sigma$ & 0.014 & 0.009 & 0.005 & 0.017 & 0.012 & 0.009 & 0.004 & 0.050 & 0.025 \\
             \cmidrule(lr){0-10}
             Baseline &  \multicolumn{10}{c}{\textbf{Textual Reference Resolution} $\Rightarrow$ \textbf{Multimodal Reference Resolution}} \\
             \multirow{2}{*}{\ w/ KWJA} & $\mu$ & \pldiff{0.319} & 0.563 & \nldiff{0.650} & \pldiff{0.328} & \pldiff{\underline{0.570}} & \pldiff{\underline{0.656}} & \pldiff{0.280} & 0.563 & \nldiff{0.672} \\
             & $\pm \sigma$ & 0.027 & 0.027 & 0.022 & 0.034 & 0.028 & 0.027 & 0.009 & 0.055 & 0.058 \\
             \multirow{2}{*}{\ w/ Ours} & $\mu$ & \pldiff{0.325} & \nldiff{0.557} & \nldiff{0.644} & 0.304 & \nldiff{0.543} & \nldiff{0.631} & \phdiff{0.305} & \pldiff{0.577} & \nldiff{0.683}  \\
             & $\pm \sigma$ & 0.014 & 0.032 & 0.026 & 0.016 & 0.033 & 0.032 & 0.025 & 0.069 & 0.058 \\
             \bottomrule
        \end{tabular}
    }
    \caption{Detail results of direct references in MRR on the J-CRe3: See the caption of Table~\ref{tab:appendix_phrase_grounding_details} for $\mu$ and $\pm \sigma$.}
    \label{tab:appendix_mmref-direct_details}
\end{table*}

\section{Ablation Study on Utterance Length}
\label{sec:appendix_ablation_window-size}
Since the evaluation of the models in Tables~\ref{tab:4_quantitative_results_coref-to-phrase_grounding}, \ref{tab:4_quantitative_results_in_mmref-indirect}, \ref{tab:appendix_phrase_grounding_details}, \ref{tab:appendix_mmref-indirect_details} and ~\ref{tab:appendix_mmref-direct_details} is based on single utterances ($\mathbf{T}$), we further conduct an ablation study on utterance length to investigate its effect on model performance.
Using longer utterances as input is expected to provide richer textual reference cues --- such as coreferences and predicate-argument structures --- than a single utterance, potentially leading to improved model performance.

\subsection{Results in Phrase Grounding}
Figure~\ref{fig:utt_grounding_pronouns} shows that increasing the input utterance length $\mathbf{T}$, which also increases the number of coreference relations contained in $\mathbf{T}$, consistently improves pronoun performance across all models.
We observe that GLIP suffers a decrease in noun performance as the utterance length increases (Figure~\ref{fig:utt_grounding_nouns}), resulting in an overall performance decrease (Figure~\ref{fig:utt_grounding_overall}).
In contrast, the MRR models based on our proposed framework (the baseline and Baseline w/ Ours) maintain stable noun performance regardless of utterance length (Figure~\ref{fig:utt_grounding_nouns}).

\subsection{Results in Multimodal Reference Resolution}
\begin{figure*}[p]
    \centering
    \begin{minipage}[t]{0.32\linewidth}
        \centering
        \includegraphics[keepaspectratio, width=\linewidth]{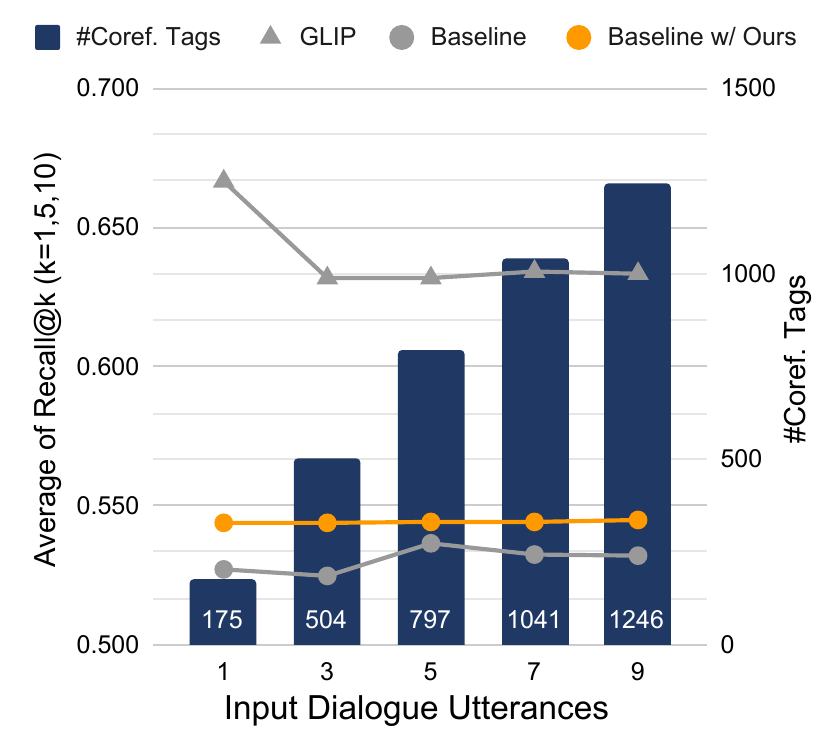}
        \subcaption{Overall}
        \label{fig:utt_grounding_overall}
    \end{minipage}
    \begin{minipage}[t]{0.32\linewidth}
        \centering
        \includegraphics[width=\linewidth]{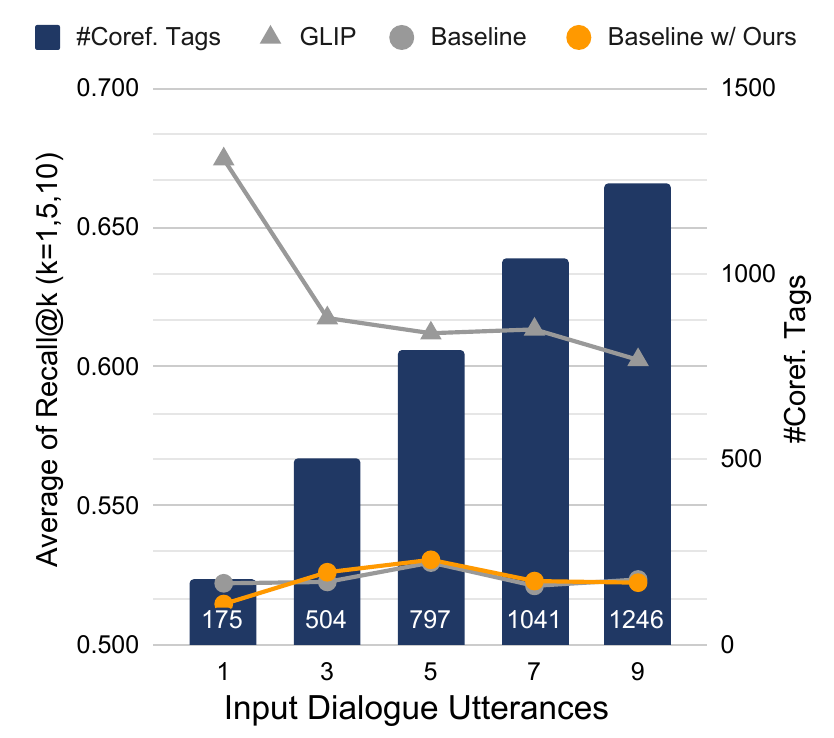}
        \subcaption{Nouns}
        \label{fig:utt_grounding_nouns}
    \end{minipage}
    \begin{minipage}[t]{0.32\linewidth}
        \centering
        \includegraphics[width=\linewidth]{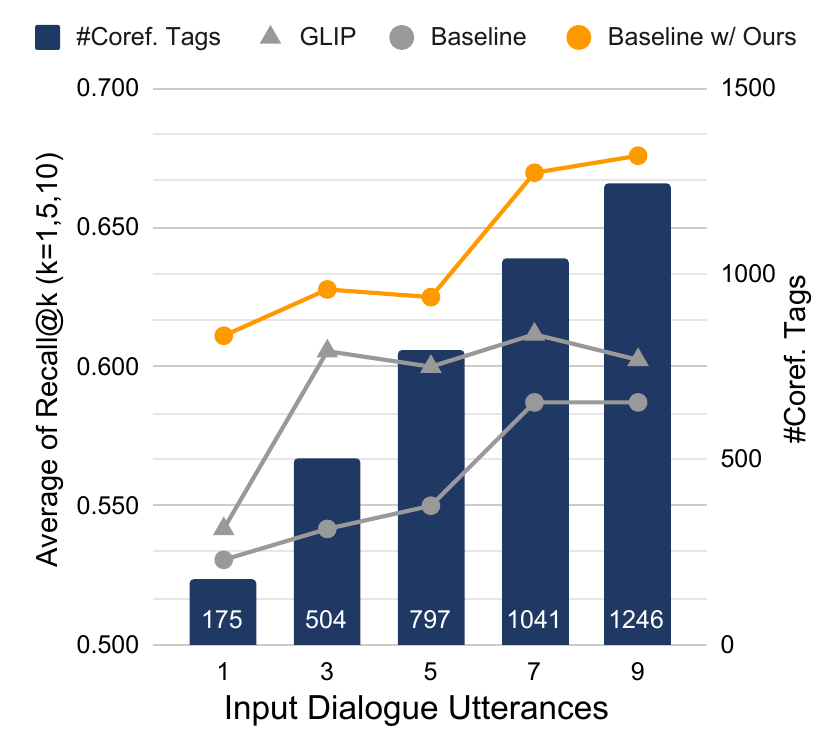}
        \subcaption{Pronouns}
        \label{fig:utt_grounding_pronouns}
    \end{minipage}
    \caption{Ablation study results on utterance length in phrase grounding: We compare GLIP, Baseline, and Baseline w/ Ours by varying the input utterance length. Changes in the average of Recall@\textit{k} ($k=\{1,5,10\}$) are shown in a range of 0.2.}

    \bigskip
    \bigskip
    
    \begin{subfigure}[b]{\linewidth}
        \begin{minipage}[b]{0.32\linewidth}
            \centering
            \includegraphics[keepaspectratio, width=\linewidth]{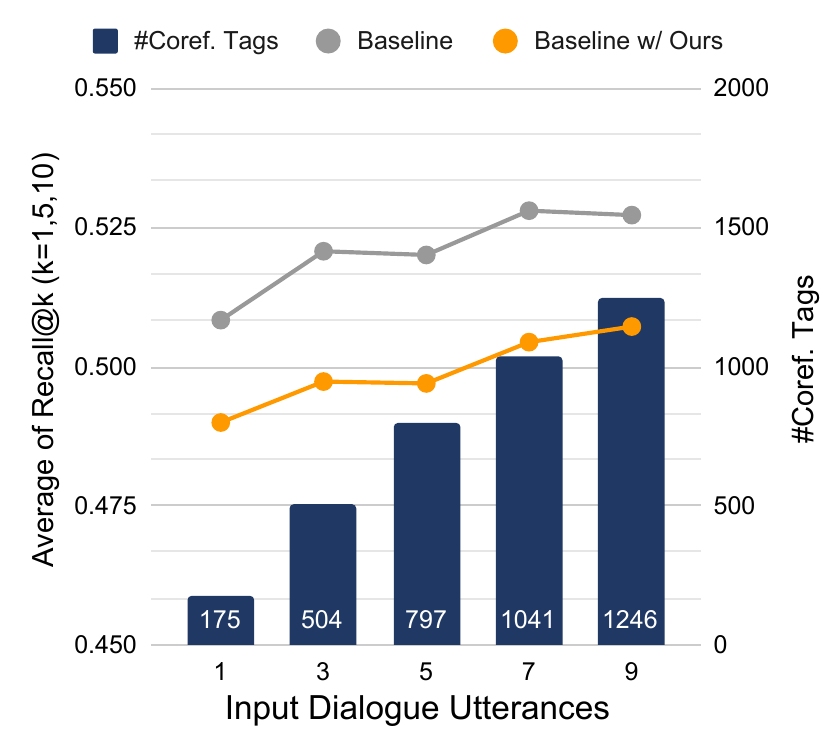}
            \subcaption{Direct references}
            \label{fig:utt_mrr_direct}
        \end{minipage}
        \begin{minipage}[b]{0.32\linewidth}
            \centering
            \includegraphics[width=\linewidth]{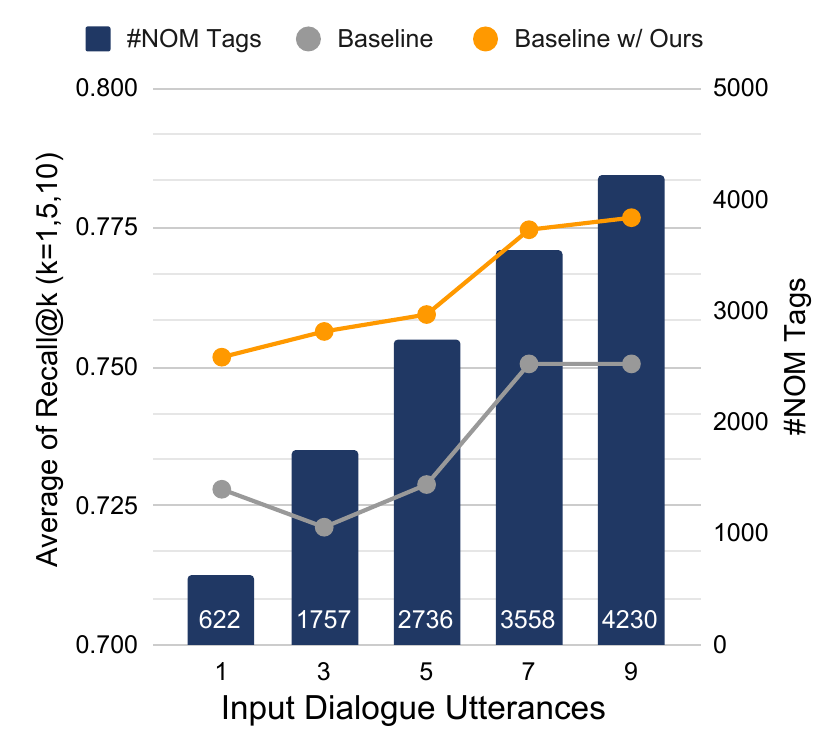}
            \subcaption{NOM}
            \label{fig:utt_mrr_nom}
        \end{minipage}
        \begin{minipage}[b]{0.32\linewidth}
            \centering
            \includegraphics[width=\linewidth]{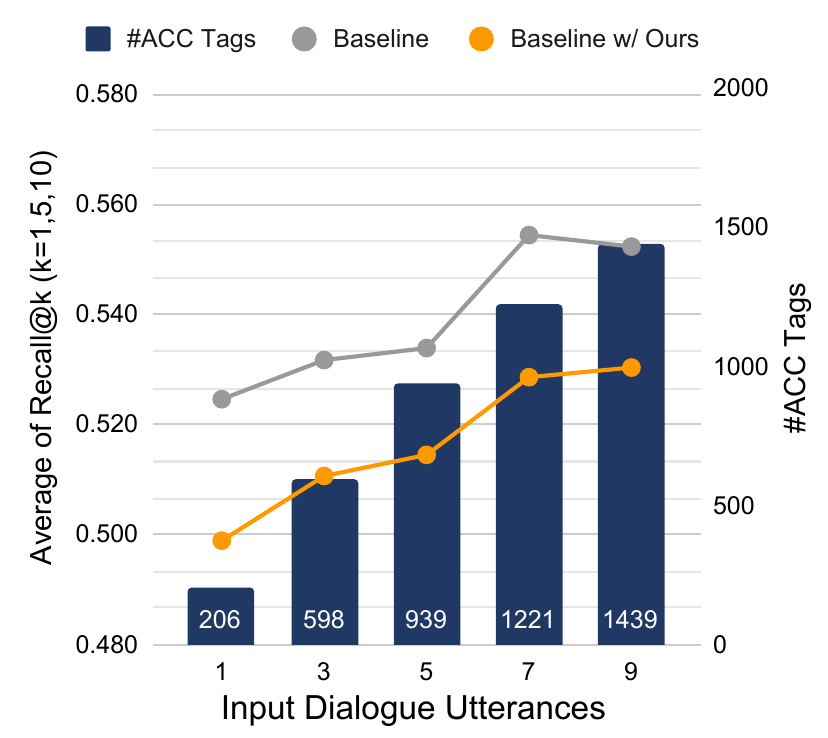}
            \subcaption{ACC}
            \label{fig:utt_mrr_acc}
        \end{minipage}
    \end{subfigure}

    \begin{subfigure}[b]{\linewidth}
        \begin{minipage}[b]{0.32\linewidth}
            \centering
            \includegraphics[keepaspectratio, width=\linewidth]{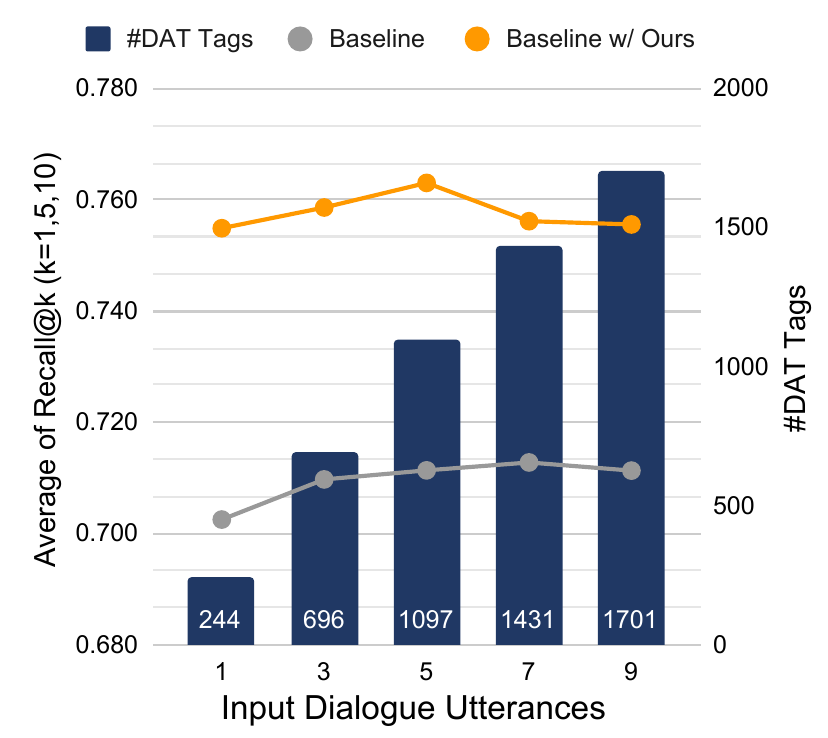}
            \subcaption{DAT}
            \label{fig:utt_mrr_dat}
        \end{minipage}
        \begin{minipage}[b]{0.32\linewidth}
            \centering
            \includegraphics[width=\linewidth]{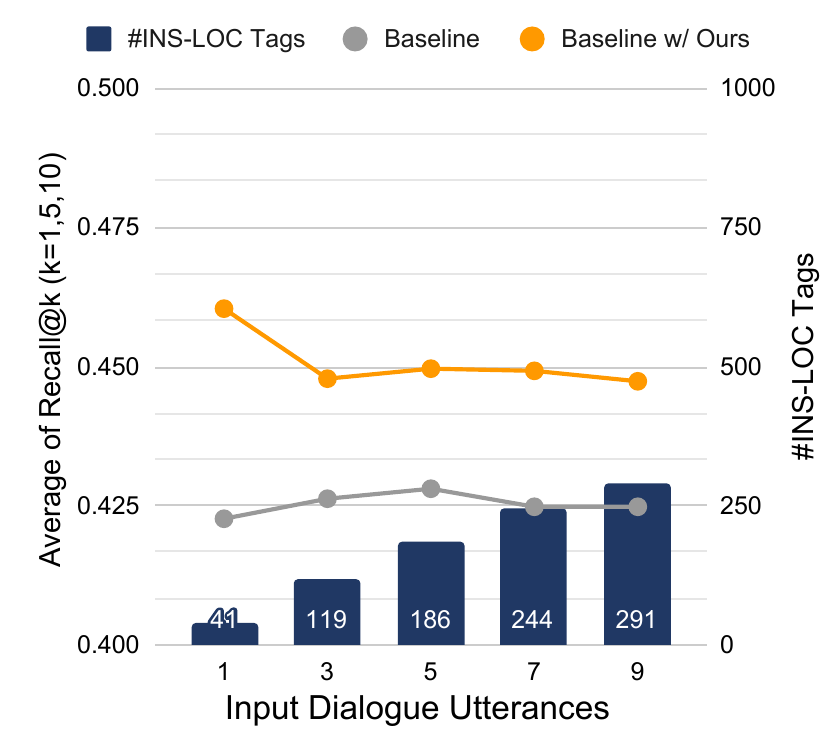}
            \subcaption{INS-LOC}
            \label{fig:utt_mrr_ins_loc}
        \end{minipage}
        \begin{minipage}[b]{0.32\linewidth}
            \centering
            \includegraphics[width=\linewidth]{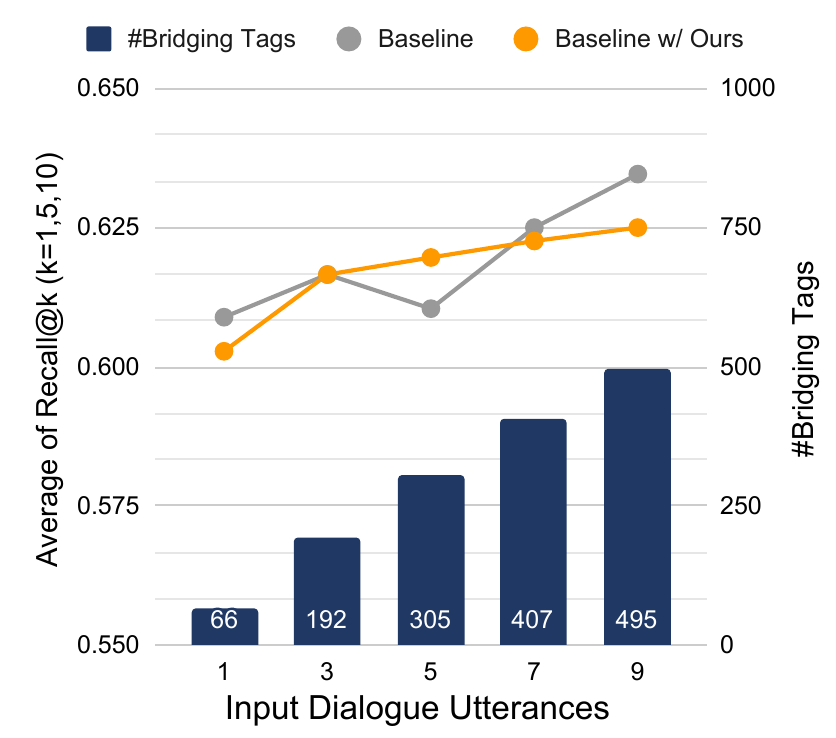}
            \subcaption{Bridging}
            \label{fig:utt_mrr_bridging}
        \end{minipage}
    \end{subfigure}
    \caption{Ablation study results on utterance length in MRR: We compare Baseline and Baseline w/ Ours by varying the input utterance length. Changes in the average of Recall@\textit{k} ($k=\{1,5,10\}$) are shown in a range of 0.1. }
    \label{fig:appendix_utterance_length}
\end{figure*}

Figure~\ref{fig:appendix_utterance_length} shows that increasing the input utterance length $\mathbf{T}$ improves performance in all models for direct references, as well as for several types of indirect references, including nominal cases (NOM), accusative cases (ACC), and bridging anaphora.
In contrast, longer utterances did not improve performance on dative cases (DAT) and instrumental and locative cases (INS-LOC).

While Figure~\ref{fig:utt_mrr_direct} shows that longer utterances lead to improved performance on direct references in MRR, Figure~\ref{fig:utt_grounding_overall} shows that no such improvement is observed in phrase grounding.
A factor contributing to the improvement observed in MRR is the increase in case relations and bridging anaphora, which are considered during evaluation.
This suggests that these types of textual reference cues can support the resolution of direct references, especially in longer utterances.
These findings are consistent with the trends observed in Figure~\ref{fig:4_ablation_mmref} and Table~\ref{tab:appendix_phrase_grounding_details}.

\section{Analysis of Confidence Score Averages}
We analyze confidence score averages to compare the baseline model with two MRR models --- one using KWJA~\cite{ueda-etal-2023-kwja} and the other using our TRR model ({\S}~\ref{sec:3_TRR_model}) --- to assess how the presence and type of TRR models affect model confidence.

\begin{figure*}[p]
    \centering
    \begin{subfigure}[b]{\linewidth}
        \begin{minipage}[b]{0.32\linewidth}
            \centering
            \includegraphics[keepaspectratio, width=\linewidth]{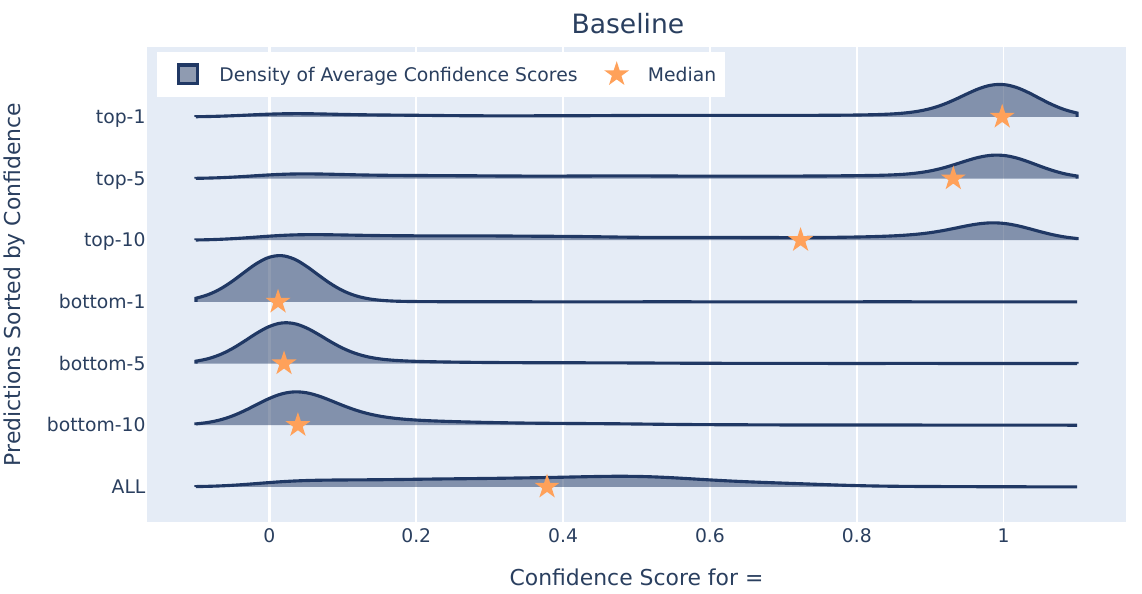}
        \end{minipage}
        \begin{minipage}[b]{0.32\linewidth}
            \centering
            \includegraphics[width=\linewidth]{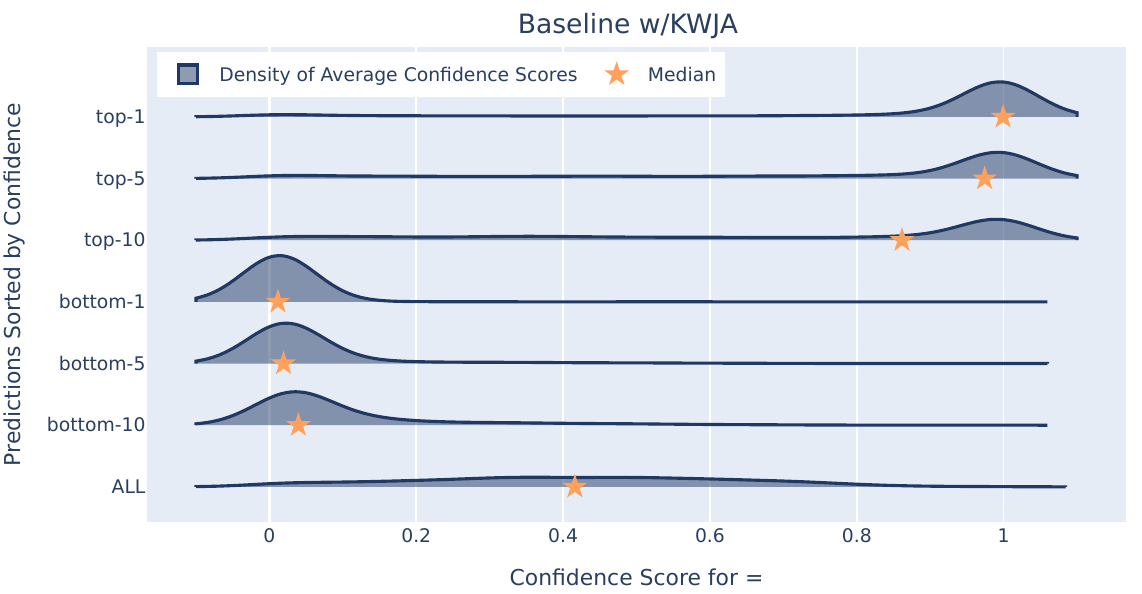}
        \end{minipage}
        \begin{minipage}[b]{0.32\linewidth}
            \centering
            \includegraphics[width=\linewidth]{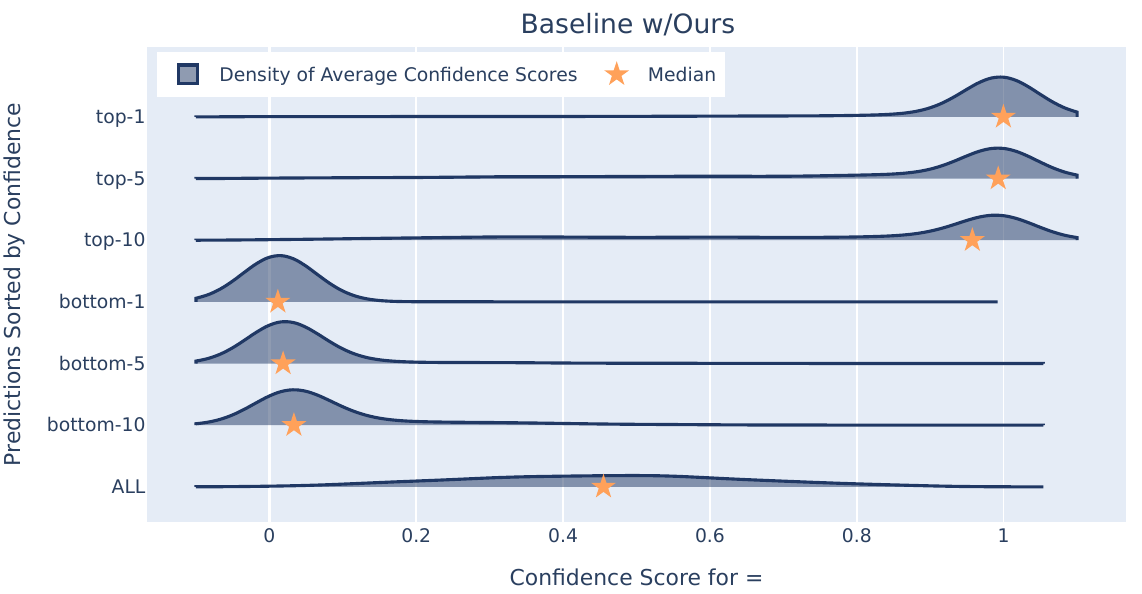}
        \end{minipage}
        \subcaption{Confidence score distribution in phrase grounding}
    \end{subfigure}
    
    \begin{subfigure}[b]{\linewidth}
        \begin{minipage}[b]{0.32\linewidth}
            \centering
            \includegraphics[keepaspectratio, width=\linewidth]{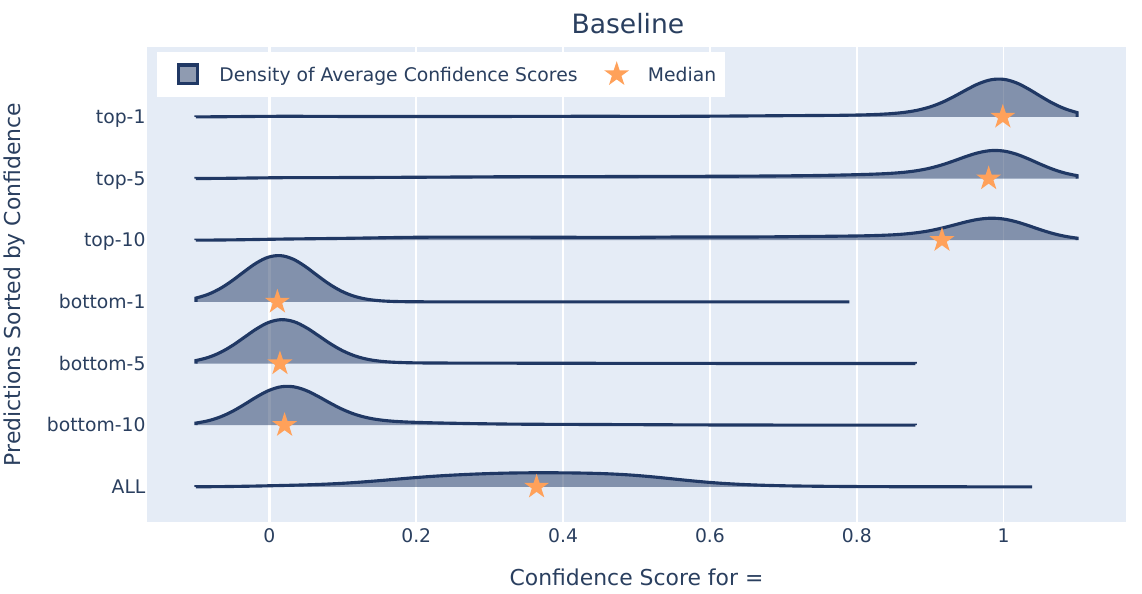}
        \end{minipage}
            \begin{minipage}[b]{0.32\linewidth}
            \centering
            \includegraphics[width=\linewidth]{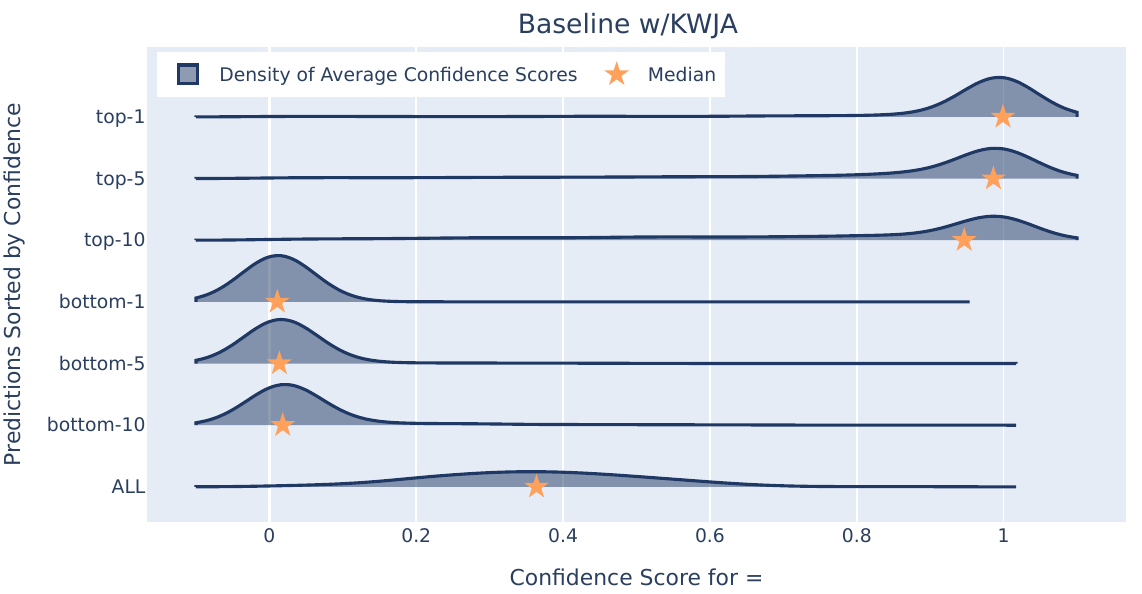}
        \end{minipage}
            \begin{minipage}[b]{0.32\linewidth}
            \centering
            \includegraphics[width=\linewidth]{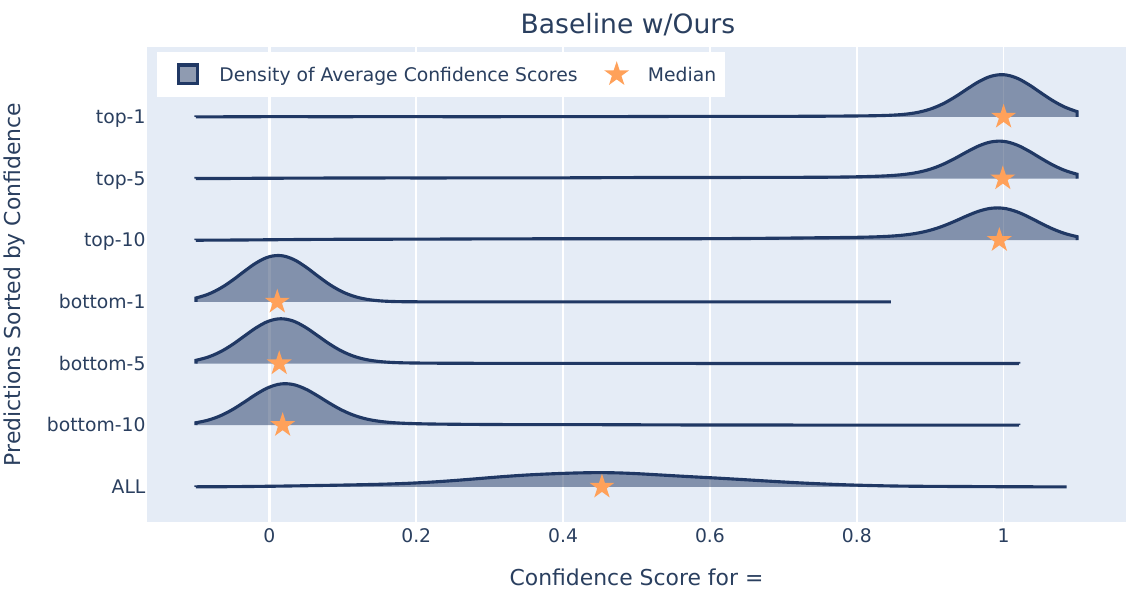}
        \end{minipage}
        \subcaption{Confidence score distribution in MRR (Direct references)}
    \end{subfigure}

    \begin{subfigure}[b]{\linewidth}
        \begin{minipage}[b]{0.32\linewidth}
            \centering
            \includegraphics[keepaspectratio, width=\linewidth]{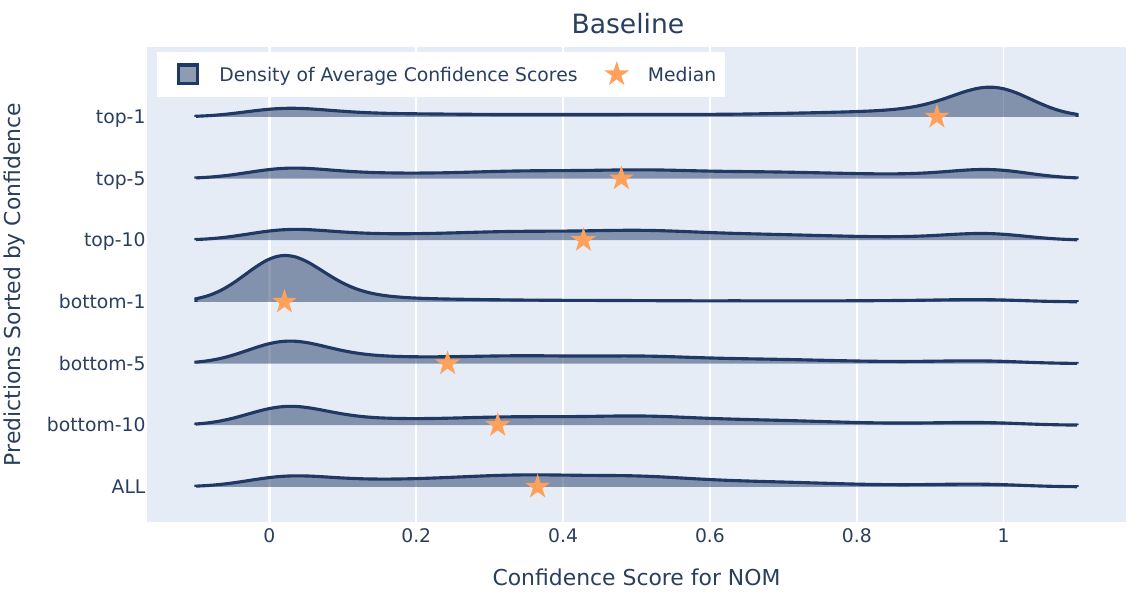}
        \end{minipage}
            \begin{minipage}[b]{0.32\linewidth}
            \centering
            \includegraphics[width=\linewidth]{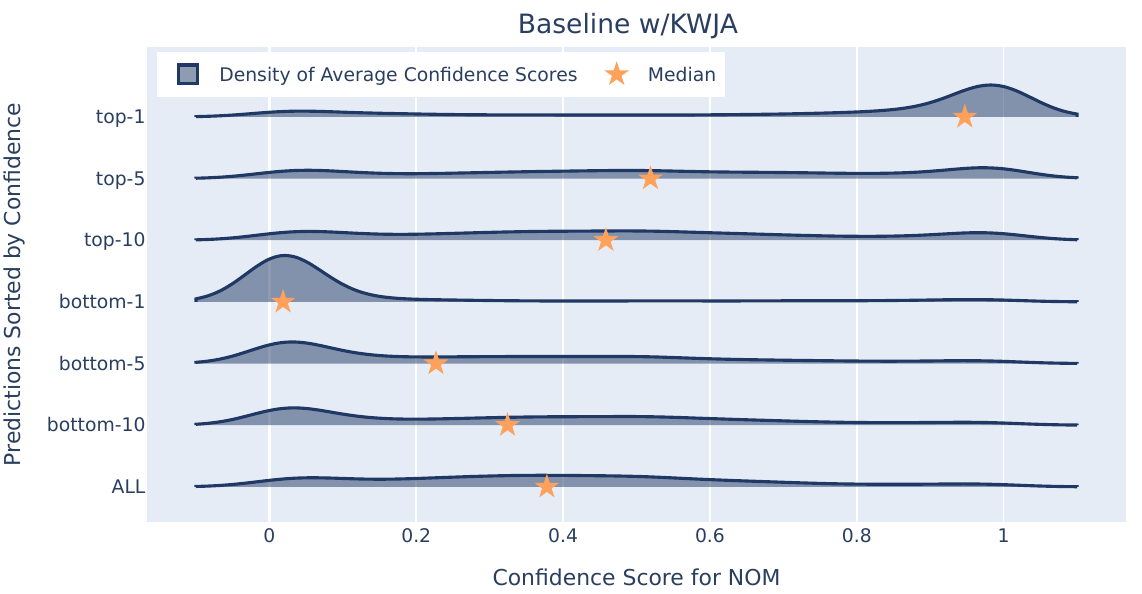}
        \end{minipage}
            \begin{minipage}[b]{0.32\linewidth}
            \centering
            \includegraphics[width=\linewidth]{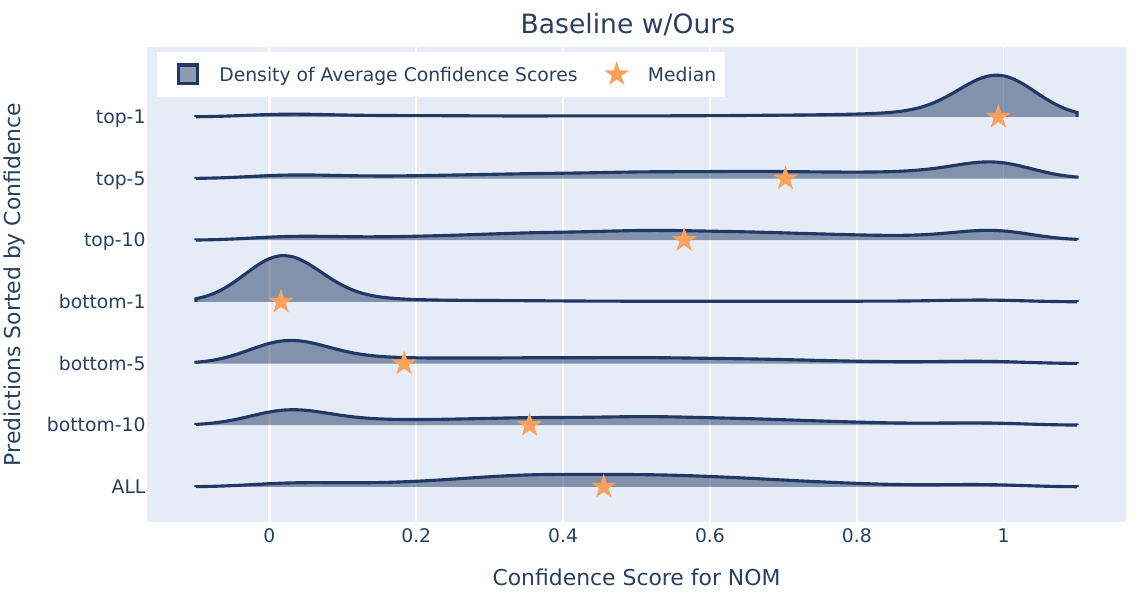}
        \end{minipage}
        \subcaption{Confidence score distribution in MRR (NOM)}
    \end{subfigure}

    \begin{subfigure}[b]{\linewidth}
        \begin{minipage}[b]{0.32\linewidth}
            \centering
            \includegraphics[keepaspectratio, width=\linewidth]{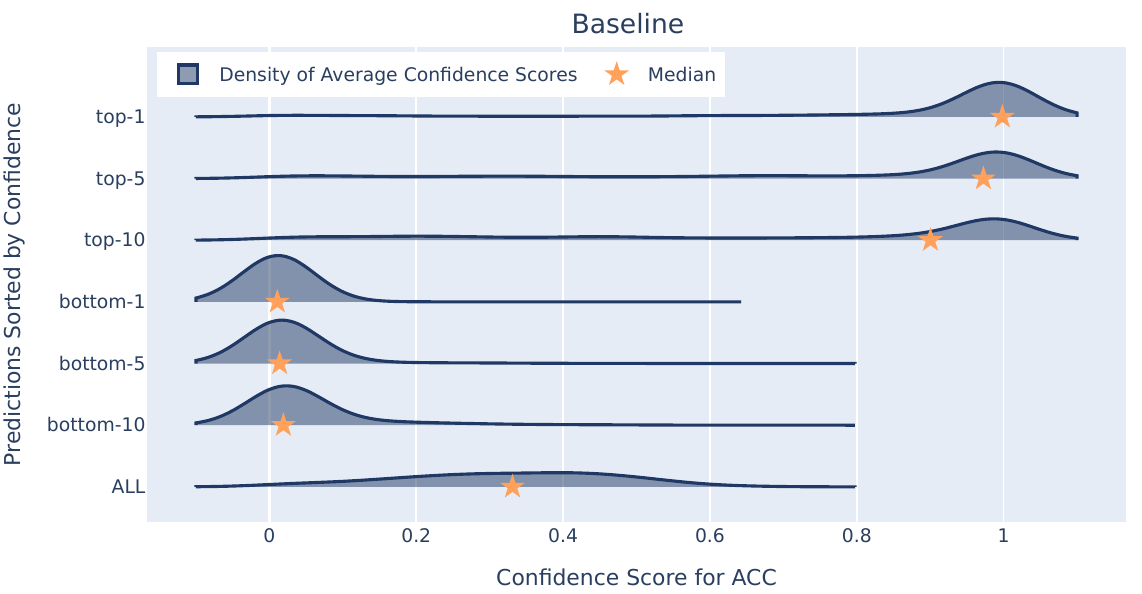}
        \end{minipage}
            \begin{minipage}[b]{0.32\linewidth}
            \centering
            \includegraphics[width=\linewidth]{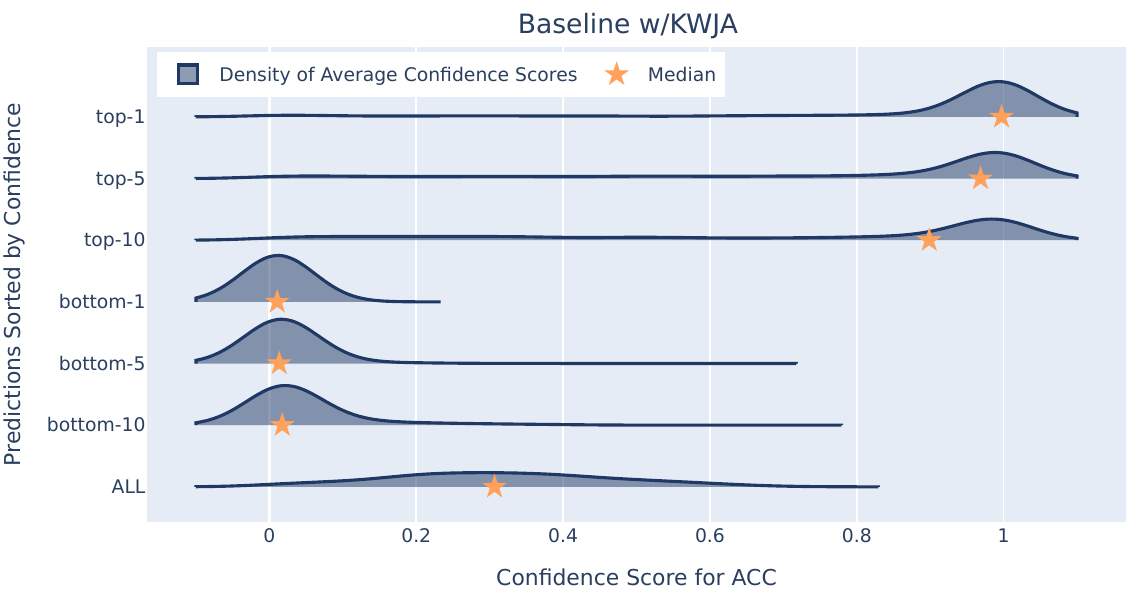}
        \end{minipage}
            \begin{minipage}[b]{0.32\linewidth}
            \centering
            \includegraphics[width=\linewidth]{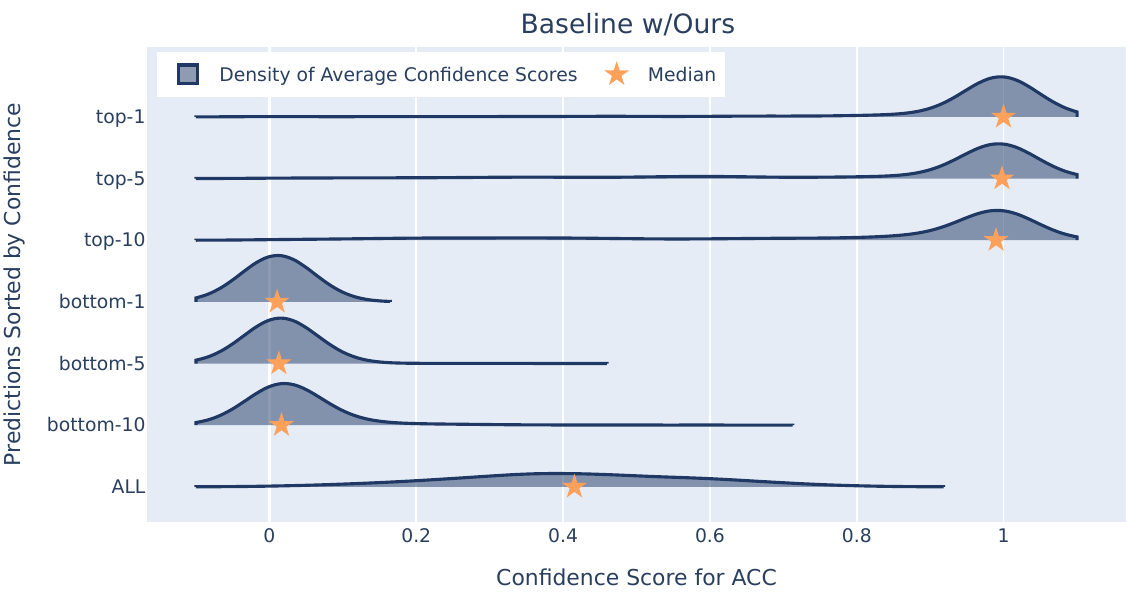}
        \end{minipage}
        \subcaption{Confidence score distribution in MRR (ACC)}
    \end{subfigure}

    \begin{subfigure}[b]{\linewidth}
        \begin{minipage}[b]{0.32\linewidth}
            \centering
            \includegraphics[keepaspectratio, width=\linewidth]{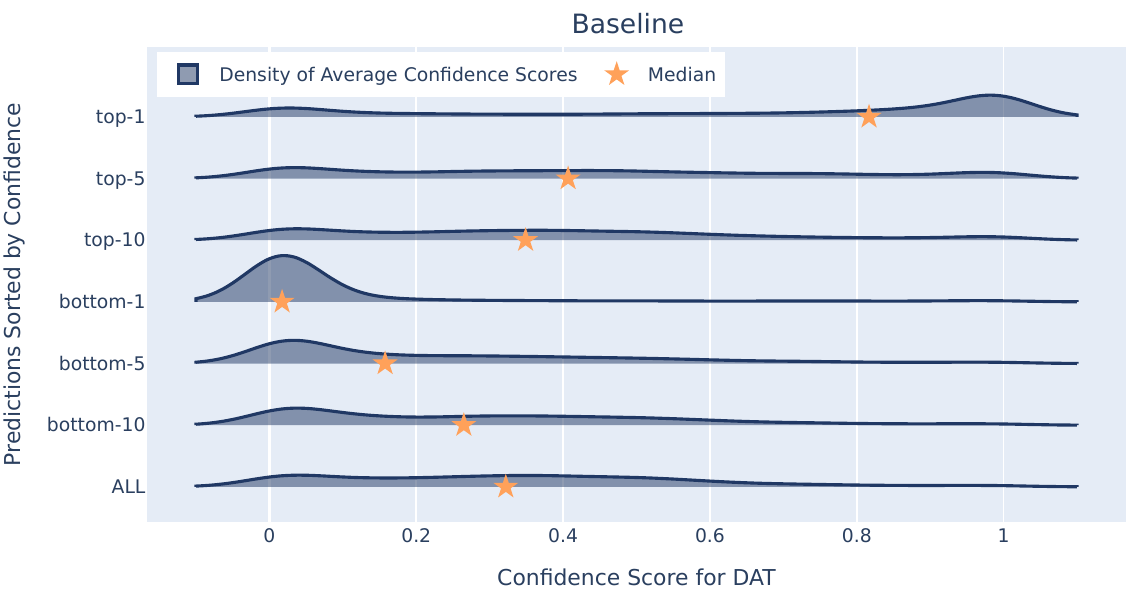}
        \end{minipage}
            \begin{minipage}[b]{0.32\linewidth}
            \centering
            \includegraphics[width=\linewidth]{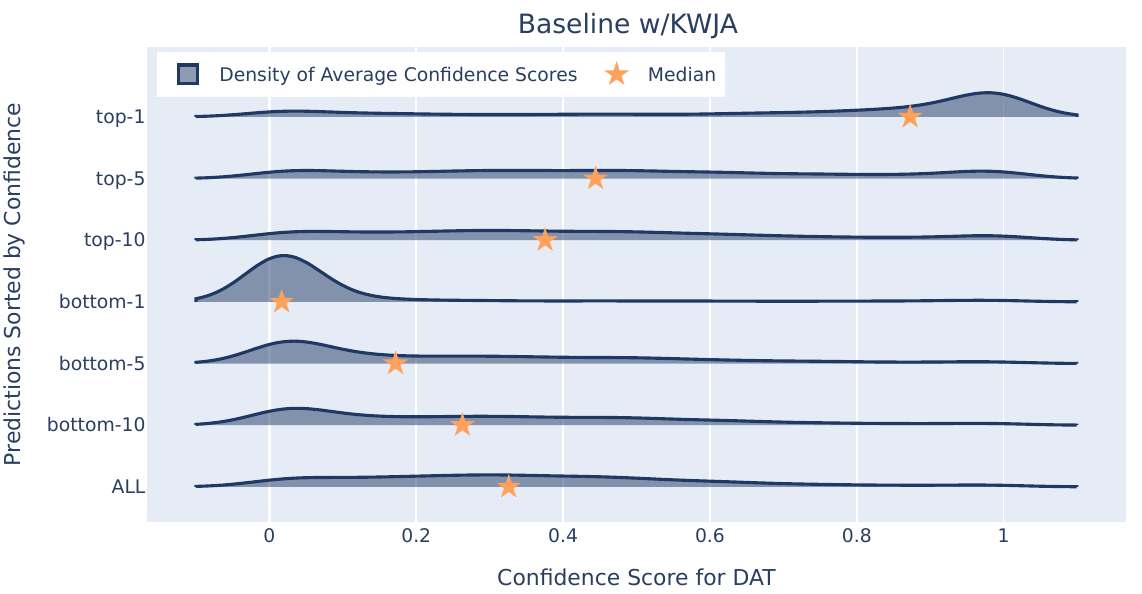}
        \end{minipage}
            \begin{minipage}[b]{0.32\linewidth}
            \centering
            \includegraphics[width=\linewidth]{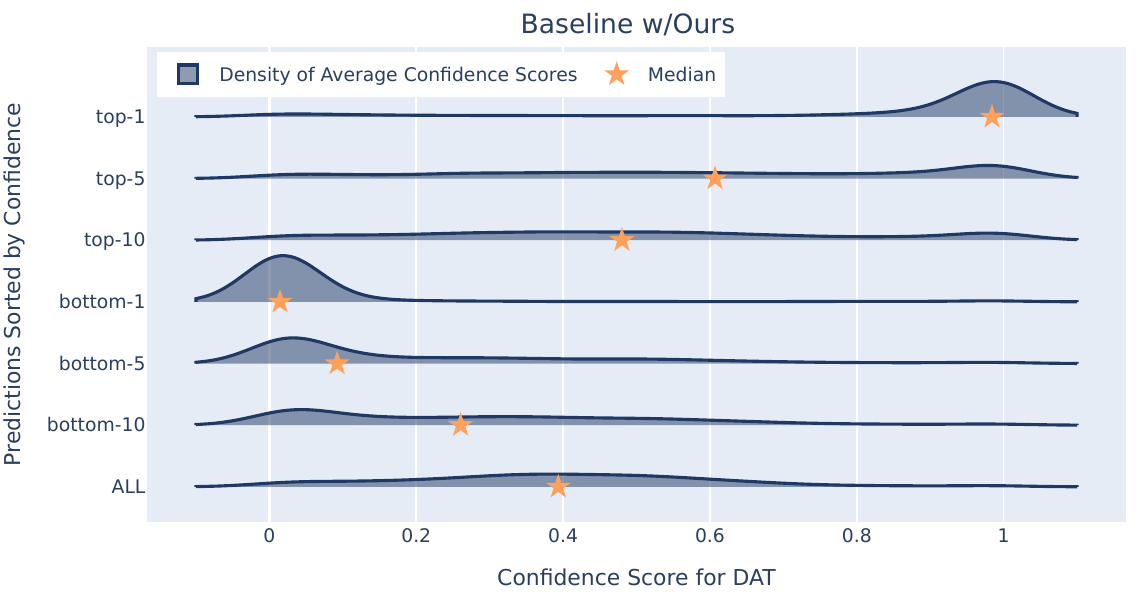}
        \end{minipage}
        \subcaption{Confidence score distribution in MRR (DAT)}
    \end{subfigure}

    \begin{subfigure}[b]{\linewidth}
        \begin{minipage}[b]{0.32\linewidth}
            \centering
            \includegraphics[keepaspectratio, width=\linewidth]{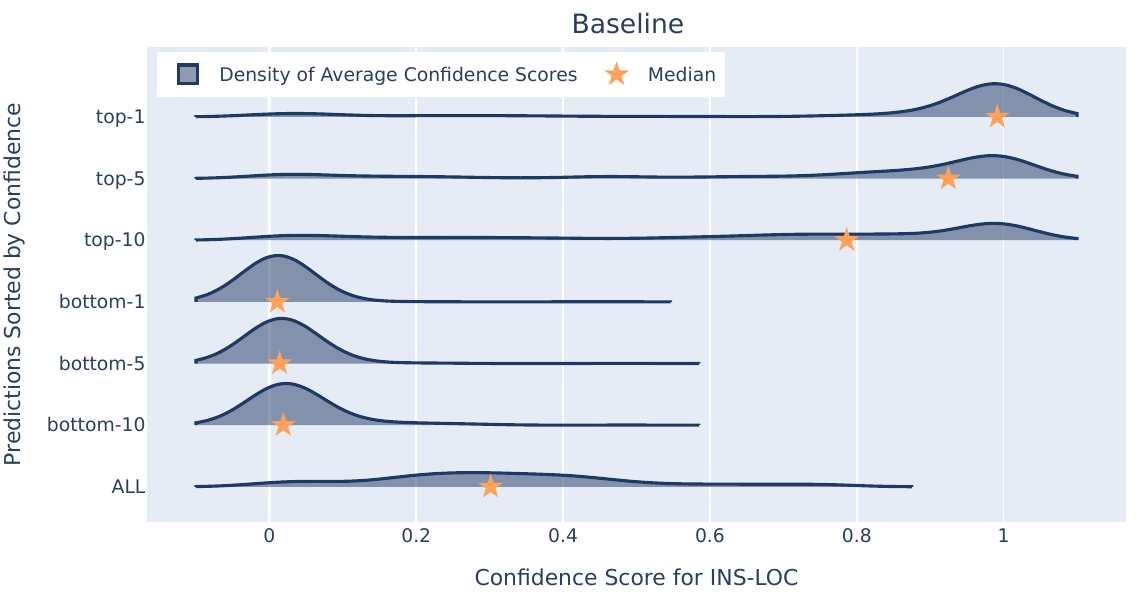}
        \end{minipage}
            \begin{minipage}[b]{0.32\linewidth}
            \centering
            \includegraphics[width=\linewidth]{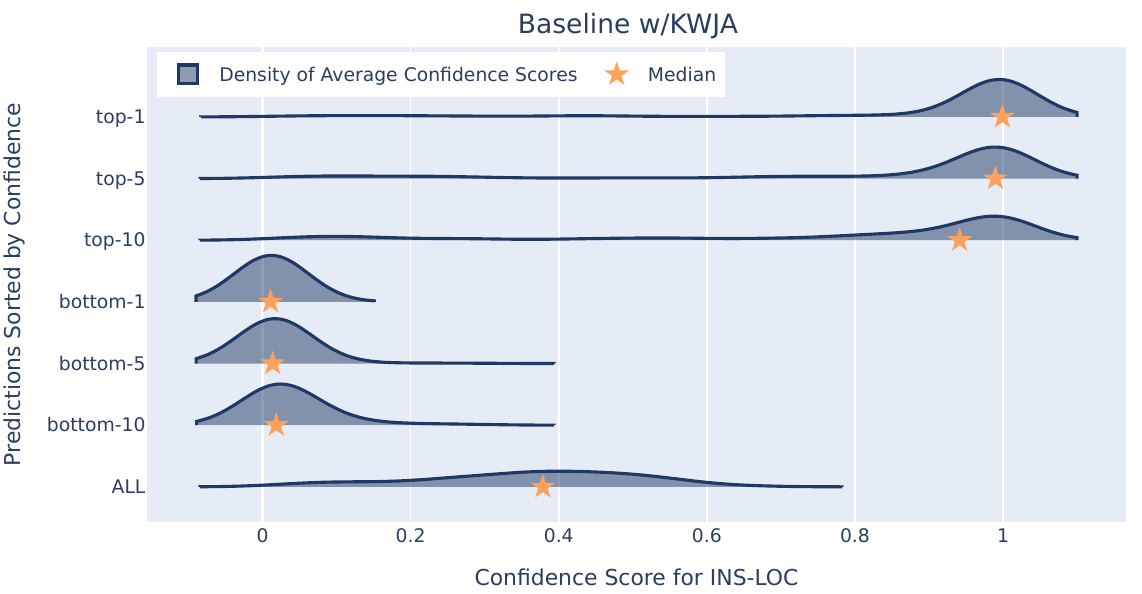}
        \end{minipage}
            \begin{minipage}[b]{0.32\linewidth}
            \centering
            \includegraphics[width=\linewidth]{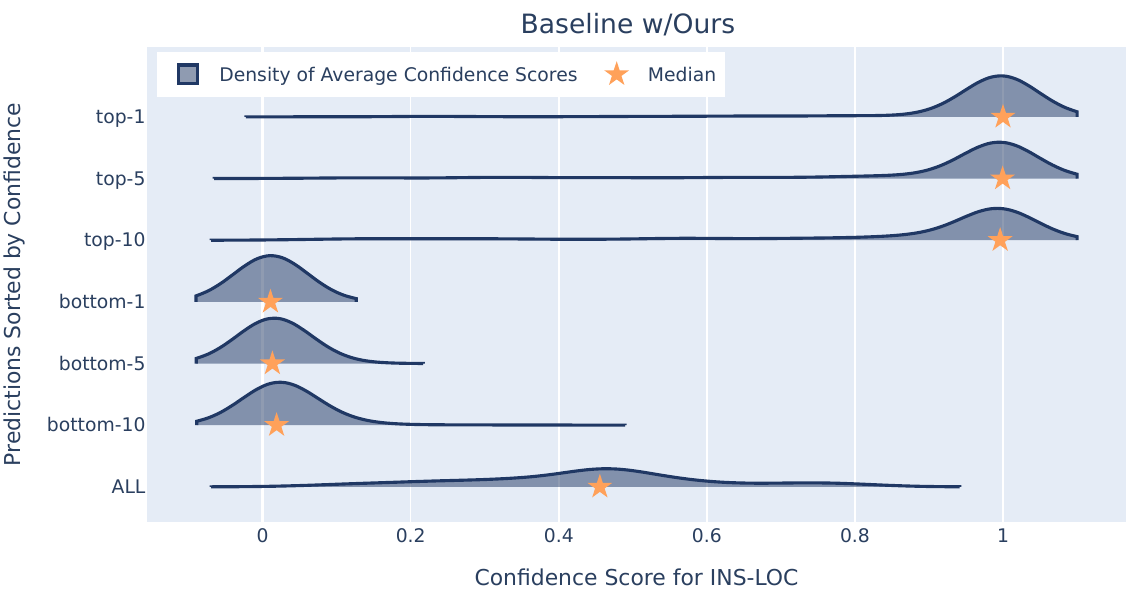}
        \end{minipage}
        \subcaption{Confidence score distribution in MRR (INS-LOC)}
    \end{subfigure}

    \begin{subfigure}[b]{\linewidth}
        \begin{minipage}[b]{0.32\linewidth}
            \centering
            \includegraphics[keepaspectratio, width=\linewidth]{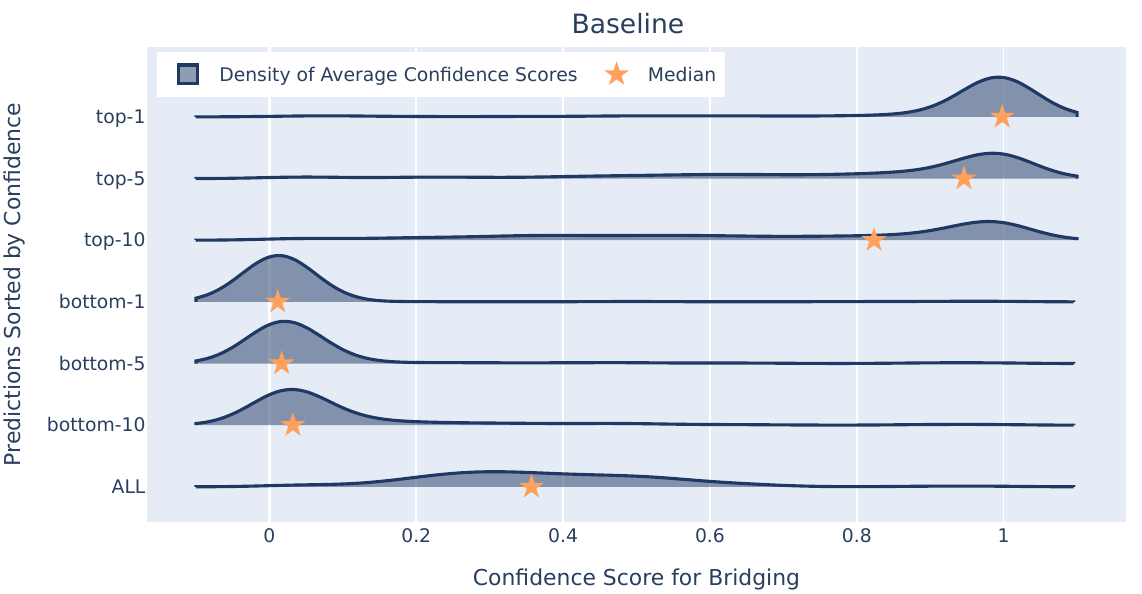}
        \end{minipage}
            \begin{minipage}[b]{0.32\linewidth}
            \centering
            \includegraphics[width=\linewidth]{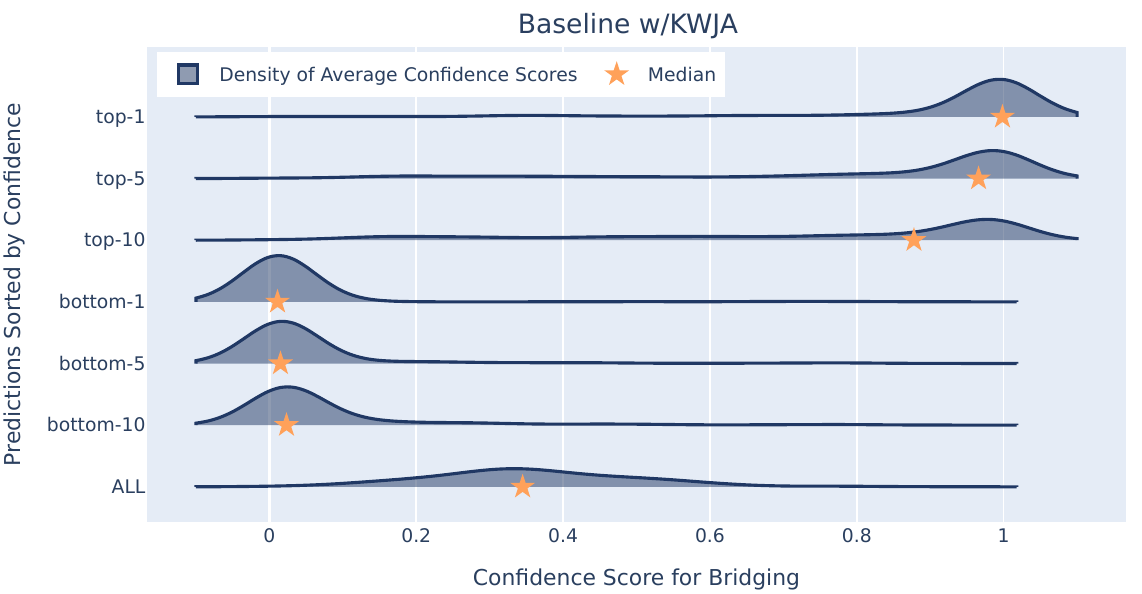}
        \end{minipage}
            \begin{minipage}[b]{0.32\linewidth}
            \centering
            \includegraphics[width=\linewidth]{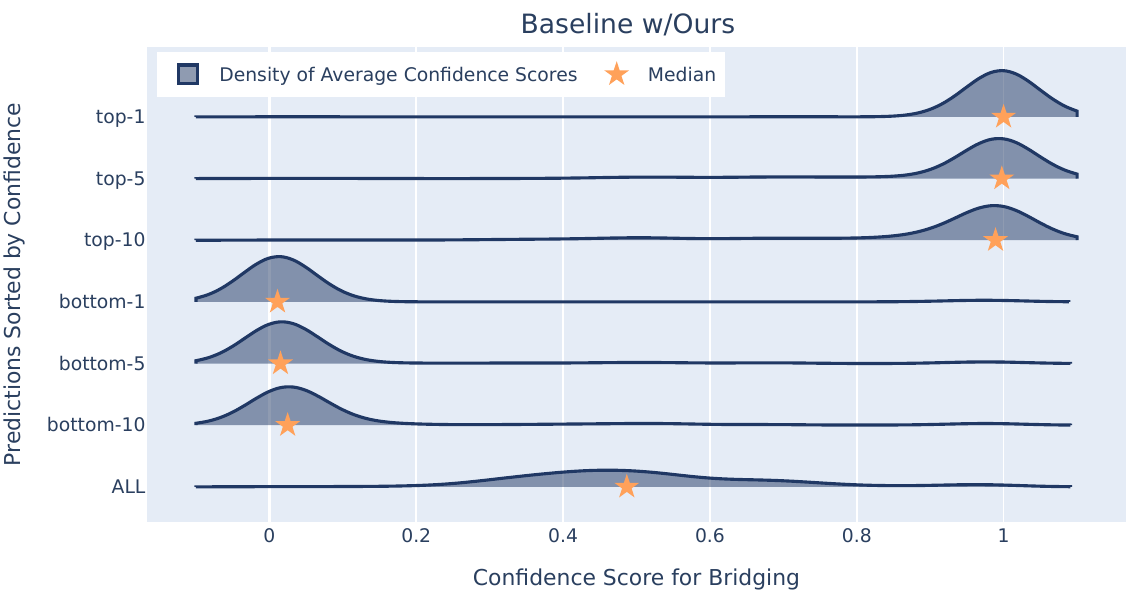}
        \end{minipage}
        \subcaption{Confidence score distribution in MRR (Bridging)}
    \end{subfigure}
    \caption{Violin plots of average confidence scores across the MRR models in phrase grounding and MRR:
    Each distribution summarizes the scores computed over Top-\textit{k}, Bottom-\textit{k} ($k=\{1,5,10\}$), and all predictions, based on model predictions from three random seeds.}
    \label{fig:appendix_confidence_distribution}
\end{figure*}

Table~\ref{fig:appendix_confidence_distribution} shows that the average confidence scores, computed over Top-\textit{k} and all predictions, exhibit a consistent rightward shift in distribution across the models --- Baseline w/ Ours, Baseline w/ KWJA, and Baseline, in that order.
This trend holds for both phrase grounding and MRR.
This quantitative result aligns with the trends observed in our qualitative analysis (Figures~\ref{fig:4_examples_grounding} and~\ref{fig:4_examples_mmref}) and suggests that incorporating textual reference, particularly in Baseline w/ Ours, which incorporates our TRR model, tends to produce higher confidence scores in predictions.

However, higher confidence scores do not always translate into better performance.
For example, as shown in Table~\ref{tab:4_quantitative_results_coref-to-phrase_grounding}, Baseline w/ Ours improves phrase grounding accuracy over the baseline, whereas Baseline w/ KWJA performs worse.
Table~\ref{tab:4_quantitative_results_in_mmref-indirect} shows that Baseline w/ Ours and Baseline w/ KWJA provide little improvement for indirect references of accusative cases (ACC) in MRR.
Collectively, these observations imply that the MRR models with TRR may suffer from overconfidence.
To address this issue, regularization strategies such as label smoothing~\cite{Szegedy_2016_CVPR} may help calibrate confidence scores.

\end{document}